%% file: arxiv_main.tex
\documentclass[10pt,twocolumn,letterpaper]{article}

\usepackage[pagenumbers]{cvpr} %

\usepackage{booktabs} %

\usepackage{algorithm}
\usepackage{algpseudocode}
\usepackage[T1]{fontenc}

\usepackage{multirow}
\usepackage{makecell}
\usepackage{amsmath}

\input{preamble}
\usepackage{wrapfig}
\usepackage{colortbl}

\usepackage{cuted}
\usepackage{balance}

\usepackage{multirow}
\usepackage{booktabs} 
\usepackage{color}
\definecolor{applegreen}{rgb}{0.55, 0.71, 0.0}
\definecolor{autumnorange}{rgb}{0.87, 0.61, 0.33}
\definecolor{moreprompt}{HTML}{C2D9FF}
\definecolor{oneprompt}{HTML}{FFE4D6}

\definecolor{heatYellow}{HTML}{FFFFB2}   %
\definecolor{heatOrange}{HTML}{FFD9B2}   %
\definecolor{heatRed}{HTML}{FFB2B2}   %

\definecolor{BrickRed}{HTML}{B5341E}
\definecolor{ForestGreen}{HTML}{309C59}

\definecolor{cvprblue}{rgb}{0.21,0.49,0.74}
\usepackage[pagebackref,breaklinks,colorlinks,allcolors=cvprblue]{hyperref}

\def\paperID{1849} %
\def\confName{CVPR}
\def\confYear{2026}

\title{GeoRelight: Learning Joint Geometrical Relighting and Reconstruction with \\ Flexible Multi-Modal Diffusion Transformers}

\author{
Yuxuan Xue\textsuperscript{1,2} \qquad
Ruofan Liang\textsuperscript{1} \qquad
Egor Zakharov\textsuperscript{1} \qquad
Timur Bagautdinov\textsuperscript{1} \qquad
Chen Cao\textsuperscript{1} \quad \\[0.25em]
Giljoo Nam\textsuperscript{1} \quad 
Shunsuke Saito\textsuperscript{1} \quad
Gerard Pons-Moll\textsuperscript{2,3} \quad
Javier Romero\textsuperscript{1} \quad\\[0.5em]
\textsuperscript{1} Codec Avatars Lab, Meta \quad
\textsuperscript{2} University of Tübingen \quad \\[0.25em]
\textsuperscript{3} Max Planck Institute for Informatics, Saarland Informatics Campus
}

\begin{document}
\maketitle

\begin{strip}
\centering
\vspace{-2.4em}
\includegraphics[width=1\linewidth]{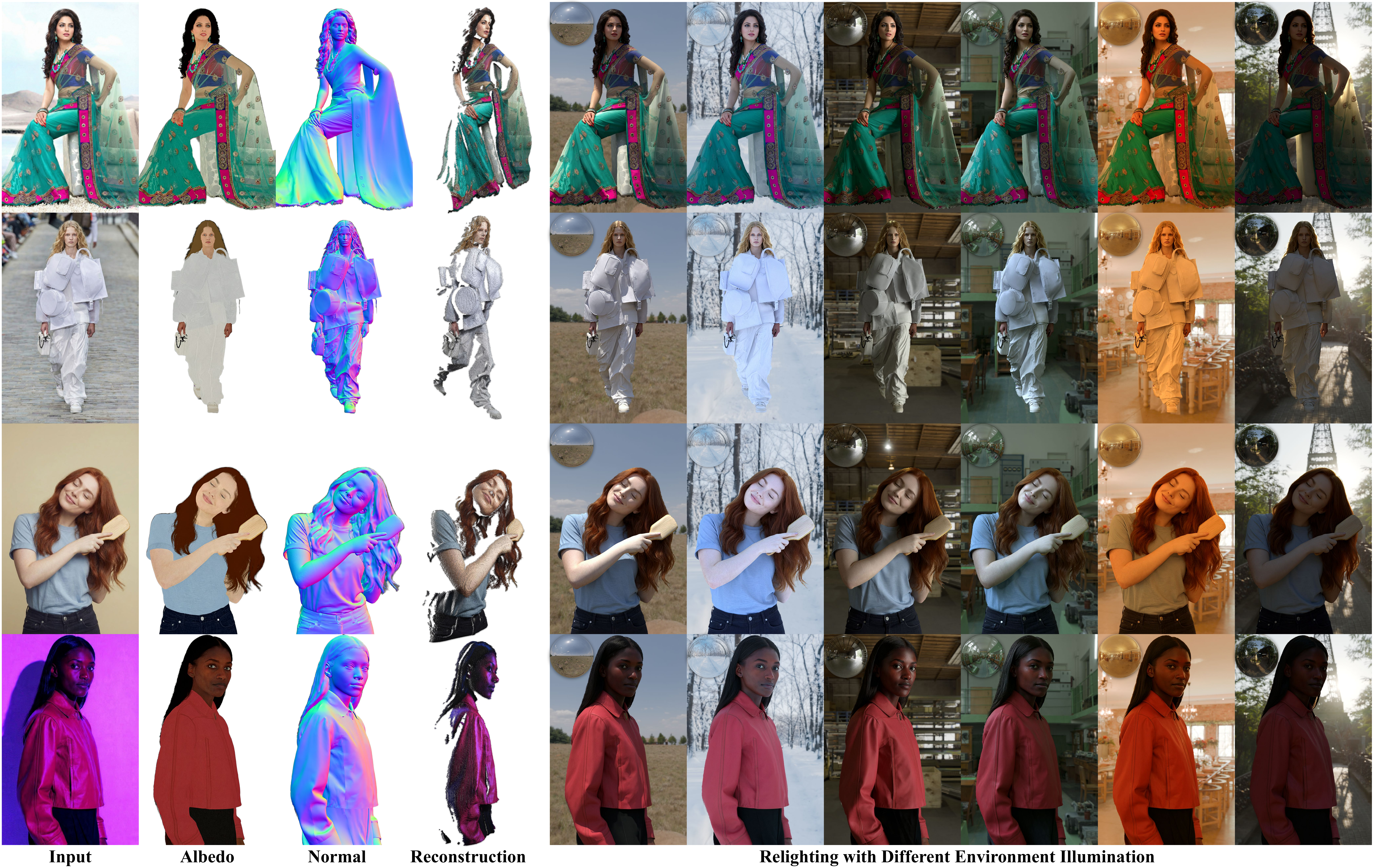}
\vspace{-1.5em}
\captionof{figure}{\textbf{GeoRelight.}
Given a monocular image (left), our framework jointly generates a relit image under novel illumination (right), 
disentangles image intrinsics like albedo (2nd column) and normals (3rd column), and extracts a fine-grained 3D pointcloud (4th column).%
}
\label{fig:teaser}
\end{strip}

\input{sec/0_abstract}    
\input{sec/1_intro}

\input{sec/2_relatedworks}

\input{sec/3_method}

\input{sec/4_experiments}

\input{sec/5_summary}

\clearpage
\input{sec/supp_body}
\appendix

\balance
{
    \small
    \bibliographystyle{ieeenat_fullname}
    \bibliography{main}
}

\end{document}

%% file: sec/0_abstract.tex
\begin{abstract}

Relighting a person from a single photo is an attractive but ill-posed task, as a 2D image ambiguously entangles 3D geometry, intrinsic appearance, and illumination. 
Current methods either use sequential pipelines that suffer from error accumulation, or they do not explicitly leverage 3D geometry during relighting, which limits physical consistency.
Since relighting and estimation of 3D geometry are mutually beneficial tasks, we propose a unified Multi-Modal Diffusion Transformer (DiT) that jointly solves for both: \textbf{GeoRelight}.
We make this possible through two key technical contributions: isotropic NDC-Orthographic Depth (iNOD), a distortion-free 3D representation compatible with latent diffusion models; and a strategic mixed-data training method that combines synthetic and auto-labeled real data. 
By solving geometry and relighting jointly, \textbf{GeoRelight} achieves state-of-the-art results in photorealistic relighting, as well as high-fidelity 3D reconstruction and intrinsic estimation from a single image.
Project page: \url{https://yuxuan-xue.com/georelight}

\end{abstract}

%% file: sec/1_intro.tex
\section{Introduction}
\label{sec:intro}
Altering illumination on a person from a single image is a core problem for computer graphics and vision~\cite{pandey2021totalrelighting, kim2024switchlight, chaturvedi2025synthlight, mei2025luxpostfacto}.
It enables enormous applications, from creative editing and computational photography to virtual reality and film production. However, the task is profoundly ill-posed. A single 2D image is an ambiguous projection of three distinct and entangled factors: the 3D geometry of the subject, their intrinsic appearance like base color, and illumination of the scene. To realistically change the light, a model must, explicitly or implicitly, disentangle these components~\cite{liang2025diffusionrenderer, zeng2024rgbx}.

Recent learning-based approaches often fall into two categories, each with a specific drawback. First, end-to-end "translator" models~\cite{chaturvedi2025synthlight, mei2025luxpostfacto, he2025unirelight, feng2025genlit} learn a direct pixel-to-pixel mapping. This approach is often limited in physical plausibility due to the absence of geometric modeling, failing to produce realistic shadows or highlights that are consistent with the subject's 3D shape. Second, and more common, are sequential pipelines~\cite{pandey2021totalrelighting, kim2024switchlight, liang2025diffusionrenderer, zeng2024rgbx}. These methods first predict a set of intermediate buffers like albedo and surface normals which are passed to a separate neural rendering module to synthesize the final image. The critical flaw in this sequential design is error accumulation~\cite{he2025unirelight}: any inaccuracies in the initial geometry estimation are baked into the buffers and cannot be corrected by the renderer.

In this paper, we argue that \textit{relighting and geometry reconstruction are not separate sequential tasks, but are mutually beneficial and must be solved jointly}. 
This synergy is two-fold. First, accurate 3D geometry is essential for the relighting model to render physically-plausible local shading and, crucially, to form correct cast shadows. Conversely, the appearance of shading in the image provides powerful shape-from-shading cues that can refine the underlying geometry estimation. We introduce \textbf{GeoRelight}, a unified generative framework that first embodies this joint approach. At its core is a Multi-Modal Diffusion Transformer (DiT)~\cite{Peebles2022DiT} that simultaneously denoises both the relit image and the 3D geometry, allowing information to flow between both tasks throughout the generation process

This joint generative approach, however, introduces two significant technical challenges. First, 3D geometry must be made compatible with a latent diffusion model that uses a pretrained 2D VAE to enable high-resolution generation. Standard representations like point maps become extremely noisy after lossy compression into VAE latent space, while traditional depth maps require normalization that anisotropically distorts the 3D shape. To solve this, we propose isotropic NDC-Orthographic Depth (iNOD) (Sec.~\ref{sec:iNOD}), a novel representation that isotropically scales the 3D shape before projection, preserving the geometry in a ``VAE-friendly'' image format without distortion. Second, this model requires data that is both photorealistic and has accurate 3D geometry labels, which is difficult to acquire. We observe that synthetic trained model is robust to produce good intrinsics and geometry. Hence use our own model to annotate pseudo labels for photorealistic images from light stage and in-the-wild data. Then, we use the strategic mixed-data training procedure (Sec.~\ref{sec:mixed_training}) where we leverage our model's flexibility to train on a mix of synthetic data (for perfect labels) and real-world data (for photorealism).

Our unified multi-modal DiT model achieves state-of-the-art results in both photorealistic relighting and high-fidelity 3D reconstruction from a single in-the-wild image. In summary, our contributions are:

\begin{itemize}
    \item  A unified Multi-Modal Diffusion Transformer framework that jointly generates relit images, intrinsic color, surface normal details, and 3D geometry.
    \item iNOD, a novel and VAE-friendly depth representation that isotropically preserves 3D geometry in latent space.
    \item A strategic mixed-data training methodology that combines synthetic data with auto-labeled real-world images to achieve both geometric accuracy and photorealism.
\end{itemize}

%% file: sec/2_relatedworks.tex
\section{Related Works}
\label{sec:related}

Image relighting and its related problems (e.g. intrinsic decomposition) have a long tradition in computer vision, going back to the work on intrinsic images by Barrow~\cite{barrow1978recovering} and the Retinex algorithm by Land~\cite{land1971lightness}. We encourage the reader to read recent surveys \cite{garces2022survey,zhu2025learning} for a wide viewpoint on intrinsic decomposition and human relighting, while we focus on the latest work on relighting in the remainder.

\subsection{Neural Rendering and Relighting}

Recent generative models have approached monocular image relighting in several ways. NeuralGaffer~\cite{jin2024neuralgaffer} treats relighting as an end-to-end 2D diffusion process, finetuning a model on a large synthetic dataset to relight any object without explicit intrinsic decomposition. IC-Light~\cite{zhang2025iclight} focuses on illumination harmonization, using diffusion to relight a foreground object to match a new background. LightLab~\cite{magar2025lightlab} enables direct manipulation of existing light sources in an image through diffusion models.

To overcome limitations from above methods, two-stage pipelines like
Total Relighting~\cite{pandey2021totalrelighting} first estimates intrinsics (e.g., albedo, normals) and then render the image with a neural shading module. This concept has been adapted in RGB-X~\cite{zeng2024rgbx} and DiffusionRenderer~\cite{liang2025diffusionrenderer}, by using separate diffusion models for forward and inverse rendering. Other methods like DiLightNet~\cite{zeng2024dilightnet} use coarse geometry estimates to create "radiance hints" that guide a ControlNet-based diffusion model. Careaga and Aksoy~\cite{careaga2025physically} leverage physical intrinsic decomposition for controllable relighting of photographs. For video, Lux Post Facto~\cite{mei2025luxpostfacto} also uses a two-stage pipeline, addressing temporal consistency by finetuning a video diffusion model on a hybrid of static OLAT (One-Light-At-a-Time) data and in-the-wild videos.

A primary drawback of sequential pipelines is their susceptibility to error accumulation, where inaccuracies from the intrinsic estimation stage are amplified by the renderer.
The advantage of estimating geometry and intrinsics was already raised by the seminal work by Barron et al.~\cite{barron2014shape} in the context of intrinsic image decomposition, without considering relighting yet.
More recently, UniRelight~\cite{he2025unirelight} introduced a general-purpose framework that jointly estimates albedo and synthesizes relit video in a single pass. By formulating this as a joint denoising problem in a video diffusion model~\cite{lu2025matrix3d, chefer2025videojam}, it implicitly reasons about scene structure. While powerful, its joint estimation is limited to albedo, leaving a critical gap in explicitly and jointly modeling the 3D geometry essential for consistent shading. %

\begin{figure}
    \centering
    \includegraphics[width=1.0\linewidth]{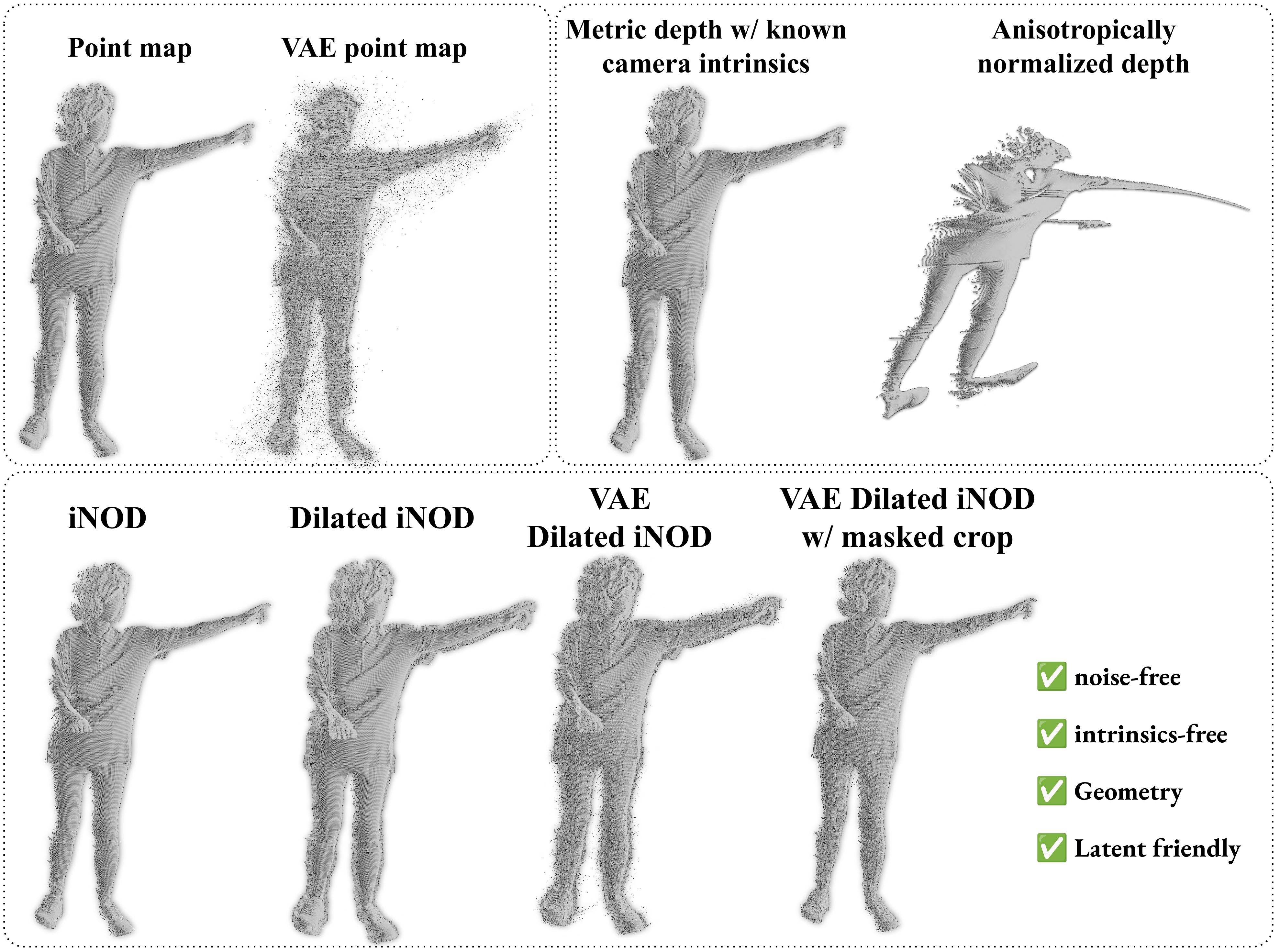}
    \vspace{-2em}
    \caption{ \textbf{iNOD: A Distortion-Free and VAE-Friendly Geometry Representation.}
   Standard Point Maps (top-left) become noisy when VAE-encoded, and anisotropically Normalized Depth (top-right) severely distorts the 3D shape.} %
    \label{fig:example_iNOD}
    \vspace{-2em}
\end{figure}

\subsection{Geometry with Latent Diffusion}
\label{sec:relatedworks_geometry}

A challenge in our joint framework is finding a 3D geometry representation compatible with high-resolution latent diffusion models that rely on pretrained VAEs to encode inputs into a latent space.
Existing representations like 3D point maps (used in ~\cite{wang2024dust3r,wang2025vggt, wang2025moge, wang2025moge2}) become noisy after being processed by the VAE (see Fig.~\ref{fig:example_iNOD}).
While other works explore adapting the VAE for point maps, they often require retraining the VAE~\cite{jiang2025geo4d}, which would prevent us from leveraging the strong generative priors of existing models.

The alternative, a 2D depth map, must be normalized to fit the VAE's $[-1, 1]$ range. Methods like Marigold~\cite{ke2025marigold, ke2023repurposing} use per-image normalization along the optical z-axis, which results in anisotropic deformations as in Fig.~\ref{fig:example_iNOD}.
Furthermore, recovering a 3D shape from such a map is impossible without knowing the camera intrinsics.
This establishes the need for a VAE-friendly representation that preserves the 3D shape isotropically, motivating our proposal (iNOD, Sec~\ref{sec:iNOD}). iNOD addresses these issues by isotropically packing a distortion-free 3D shape into an image format that is both robust to VAE compression and requires no camera intrinsics at inference time.

%% file: sec/3_method.tex
\section{Methods}
\label{sec:method}
\begin{figure*}
    \centering
    \includegraphics[width=1.0\linewidth]{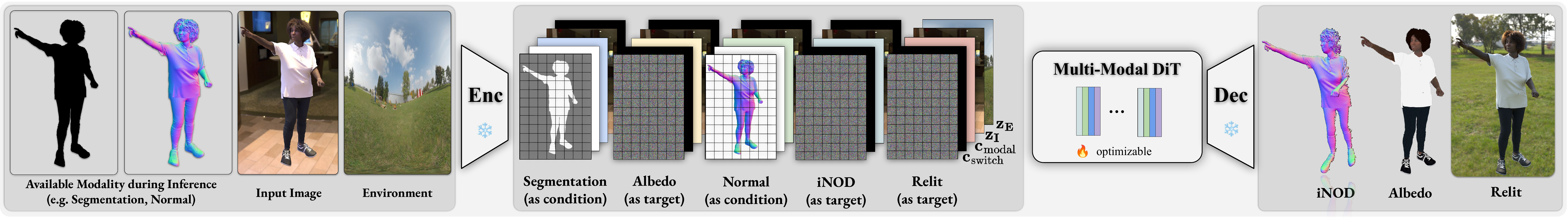}
    \vspace{-2em}
        \caption{\textbf{The GeoRelight Pipeline.}
        GeoRelight processes up to five target modalities, using $\mathbf{c}_{\text{switch}}$ to signal which ones are targets and conditions (the figure shows one specific usecase). It is guided by a global image condition $\mathbf{z}^\mathbf{I}$ and a specific illumination condition $\mathbf{z}^\mathbf{E}$. 
}
    \label{fig:pipeline}
    \vspace{-1.5em}
\end{figure*}
Given a single image of a person captured under unknown illumination, our goal is to jointly relight the person with a desired target environment $\mathbf{E}$ while also recovering their intrinsic albedo $\mathbf{a}$ and 3D geometry $\mathbf{g}$. We formulate this joint relighting and reconstruction problem as a conditional multi-modal generation task.

To enable high-resolution generation and leverage strong generative priors, we build our framework on a Video Latent Diffusion Transformer (DiT). This architectural choice is powerful, but it introduces three core technical challenges.
First, the model must be flexible enough to generate and be conditioned on multiple, distinct modalities at once; we address this with our Flexible Multi-Modal Diffusion design (Sec.~\ref{sec:flexible_mm_diffusion}). Second, we must represent 3D geometry in a ``VAE-friendly'' way that avoids the distortion of standard depth maps; we propose a novel representation, isotropic Normalized Orthographic Depth (iNOD) (Sec.~\ref{sec:iNOD}) to solve this. Finally, we must learn photorealism when real-world images lack ground-truth geometry, which we solve with a Strategic Mixed-Data Training procedure (Sec.~\ref{sec:mixed_training}).

\subsection{Flexible Multi-Modal Diffusion}
\label{sec:flexible_mm_diffusion}

Our goal of jointly modeling geometry and relighting requires an architecture that can process multiple distinct modalities at once (e.g., images, albedo, surface normal, and geometry). A Diffusion Transformer (DiT) is a natural choice, as its dense self-attention mechanism allows for intensive information exchange between all modalities simultaneously. Furthermore, operating in the latent space of a pretrained VAE enables high-resolution generation while preserving fine-grained details.

We adapt pre-trained video DiT that denoises latent $\mathbf{z}{\in}\mathbb{R}^{T\times H\times W\times C}$, where $T$ is temporal length, $H$ and $W$ are spatial resolution, $C$ is latent channel dimension. Inspired by~\cite{he2025unirelight}, we adapt it with a simple but effective modification: \textit{we repurpose the temporal dimension $T$ as a modality dimension $M$}. This allows us to stack all our target modalities (albedo, normal, geometry, segmentation, relit image) and process them in parallel as if they were frames in a video.

This ``modality-as-time'' design requires specific conditioning. We use a \textit{learnable modality type embedding} $\mathbf{c}_\text{modal}{\in}\mathbb{R}^{M\times C_\text{type}}$ to distinguish which ``frame'' (modality) is being processed. Before being passed to DiT blocks, each modality's embedding is broadcast to $\mathbb{R}^{H\times W \times C_\text{type}}$ and concatenated channel-wise to its corresponding latent. 
Furthermore, to ensure that a pixel at position (x,y) in one modality (e.g. relit image) can be spatially correlated with the same pixel in others (e.g. normal map) in the attention mechanism, we use a \textit{shared spatial positional embedding} to replace original temporal 3D RoPE~\cite{su2023rope} across all modalities, making the modality's own position invariant. More specifically, we apply the same 2D RoPE to all modalities.

Finally, the key to our framework's flexibility is the modality switch mask $\mathbf{c}_{\text{switch}} {\in} \mathbb{R}^{H\times W\times 1}$. This binary mask, concatenated to each modality, signals whether it is a ``clear'' input condition (use 1) or a ``noisy'' target to be generated (use 0). This switch implicitly tells the model to either condition on this modality or to generate it. This mechanism is the core enabler for our strategic mixed-data training (Sec.~\ref{sec:mixed_training}) and allows for flexible, powerful control during inference, as demonstrated in our ablations.

\subsection{Joint Relighting and Reconstruction}

\begin{figure*}
    \centering
    \includegraphics[width=1.0\linewidth]{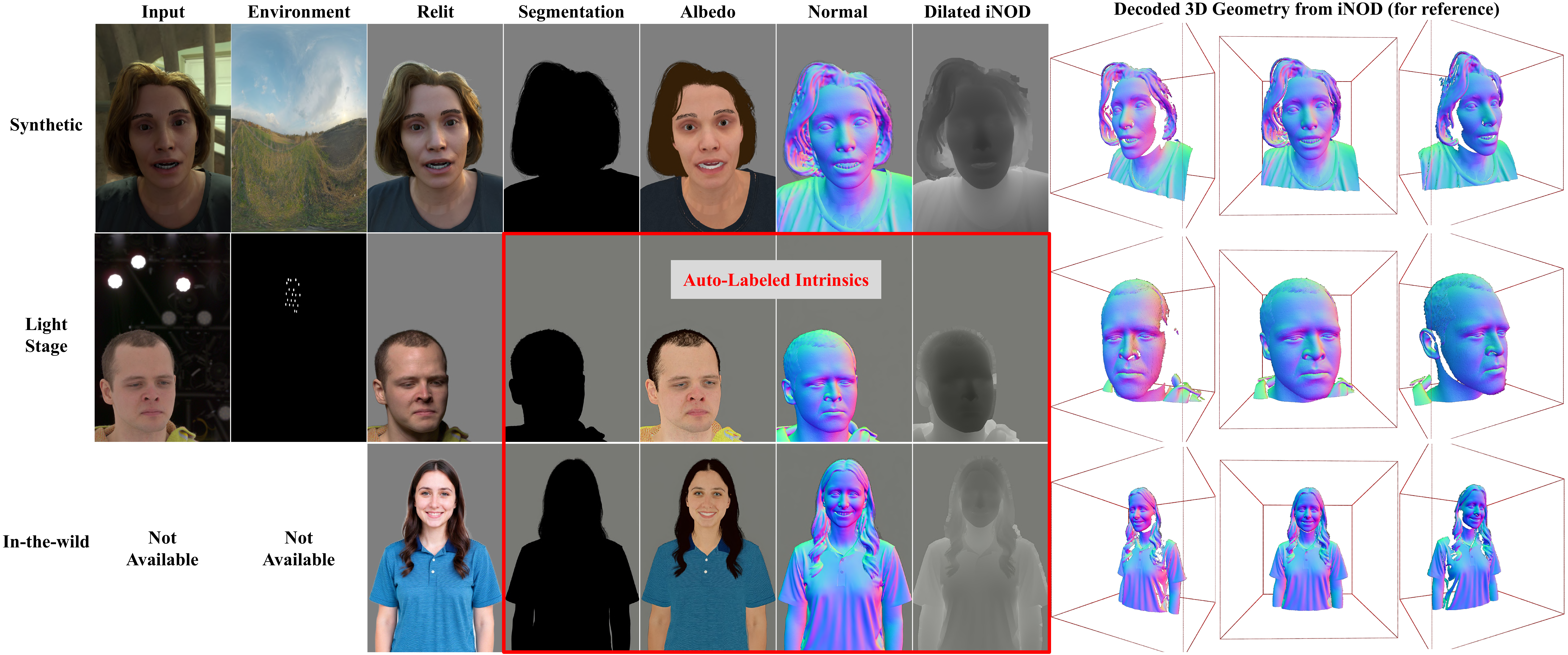}
    \vspace{-2em}
    \caption{\textbf{Our Strategic Mixed-Data Training Sources.}
    We combine (a) fully-labeled Synthetic data, (b) Light Stage data with paired lighting, and (c) In-the-wild data. We use our synthetic data pre-trained model to auto-label intrinsics for (b) and (c).} %
    \vspace{-1.5em}
    \label{fig:trainingdata_example}
\end{figure*}

As illustrated in Figure~\ref{fig:pipeline}, our model is designed to jointly denoise five target latent modalities from a single input image $\mathbf{I}$: intrinsic albedo $\mathbf{z}^{\mathbf{a}}$, a segmentation mask $\mathbf{z}^{\mathbf{s}}$, surface normals $\mathbf{z}^{\mathbf{n}}$, geometric shape $\mathbf{z}^{\mathbf{g}}$, and final relit image $\mathbf{z}^{\mathbf{I}_{\mathbf{E}}}$.

To guide this complex multi-modal generation, we introduce two distinct types of conditioning. We use the original input image as \textit{global condition}. This image contains the core information about the subject's identity, shape, and texture that all target modalities must share. Unlike UniRelight~\cite{he2025unirelight}, which treats the input as just another modality to denoise, we concatenate latent $\mathbf{z}^{\mathbf{I}}$ channel-wise to all five target modalities $\mathbf{z}^{m}$, ensuring its information is present in every step of the joint denoising process uniformly.

The second is the illumination condition E, which is required only for the relit image $\mathbf{z}^{\mathbf{I}_{\mathbf{E}}}$. A key challenge is that $\mathbf{E}$ is a High Dynamic Range (HDR) map, which the pretrained LDR-based VAE cannot encode faithfully. To solve this, we follow the established practice of~\cite{jin2024neuralgaffer, liang2025diffusionrenderer} and decompose the HDR map into three LDR-compatible representations: a Reinhard tonemapped~\cite{reinhard2005tonemapping} panorama $\mathbf{E}_{\text{ldr}}$, a normalized log-intensity map $\mathbf{E}_{\text{log}}$, and a directional map $\mathbf{E}_{\text{dir}}$. These are encoded into a joint lighting latent $\mathbf{z}^{\mathbf{E}}$. This latent is concatenated only to the relit image latent $\mathbf{z}^{\mathbf{I}_{\mathbf{E}}}$, while zero-padded placeholders are used for all other modalities to maintain a consistent tensor shape.

Thus, the full input for any single modality m in our DiT is a stack containing its noisy latent $\mathbf{z}^{m}_{\tau}$, the global condition $\mathbf{z}^{\mathbf{I}}$ , its specific illumination condition (either $\mathbf{z}^{\mathbf{E}}$ or zeros), and our flexible conditioning embeddings $\mathbf{c}_{\text{modal}}$ and $\mathbf{c}_{\text{switch}}$. The model is then optimized using a denoising score matching objective~\cite{karras2022edm}. At inference, this design allows us to either generate all modalities from pure noise or, by using the $\mathbf{c}_{\text{switch}}$, provide any combination of clear modalities to guide the generation of the others.

\subsection{Isotropic Normalized Orthographic Depth}
\label{sec:iNOD}
As we established in Sec.~\ref{sec:relatedworks_geometry} and~\ref{sec:flexible_mm_diffusion}, a VAE-friendly geometry representation is critical for our joint model. 
Standard point maps are corrupted by VAE encoding and normalized depth is anisotropic  (Fig.~\ref{fig:example_iNOD}).
To solve this, we propose isotropic Normalized Orthographic Depth (iNOD), a simple yet effective representation that is dense, VAE-friendly, and perfectly preserves relative 3D geometry.

The creation of an iNOD is straightforward. Given ground-truth metric depth and camera intrinsics (available during synthetic data creation), we first unproject the pixels to acquire a metric 3D point cloud. The key step is our isotropic 3D normalization: instead of scaling only the z-axis, we scale the entire 3D geometry to fit within a [-1, 1] bounding box based on its longest edge. This operation discards absolute scale but maintains the relative 3D geometry and aspect ratio of the subject. 
From this normalized 3D shape, we orthographically project it onto the XY-plane by simply taking the z-value of each point. This final orthographic depth map is naturally in the [-1, 1] range required by the pretrained VAE. We then assign this z-value back to the corresponding pixel to create a dense, pixel-aligned 2D map that represents a distortion-free 3D shape.

This iNOD representation has all the properties we require. It is a VAE-friendly 2D image with 1-DoF that eliminates the noise artifacts seen in point map encoding with 3-DoF. 
Most importantly, because the normalization is isotropic, the 3D non-distorted shape can be recovered with a simple orthographic unprojection, eliminating the need for camera intrinsics during inference, which is a major advantage over standard depth maps. We find that VAE compression can create boundary artifacts due to nearest neighbor sampling, so we apply a simple dilation to the iNOD foreground, which results in a clean and robust geometry boundary after decoding shown in Fig.~\ref{fig:example_iNOD}. Please refer to our supplementary for detailed instructions on creating iNOD.

\subsection{Strategic Mixed Post-Training}
\label{sec:mixed_training}

A fundamental challenge for using synthetic data is the ``synthetic-real gap''. Synthetic data provides perfect ground-truth labels but it may result in a model that achieves less photorealism. Conversely, real-world images are photorealistic but lack the ground-truth intrinsics and geometry labels, and are extremely complicated to acquire.

To solve this, we first create a hybrid dataset. In our observation, synthetic trained model achieves good intrinsics and geometry recovery. Hence, we leverage our initially trained synthetic model to ``auto-label''~\cite{liang2025diffusionrenderer} our real-world data sources: a high-quality light stage data (\texttt{Dome}) as in~\cite{martinez2024castudio, wang2025fullbodyavatar} and a large-scale in-the-wild dataset (\texttt{ITW}) from~\cite{li2025cosmicman, laion2024relaion}. This step provides us with pseudo-ground-truth intrinsics ($\mathbf{z}^{\mathbf{a}}$, $\mathbf{z}^{\mathbf{n}}$, $\mathbf{z}^{\mathbf{g}}$, $\mathbf{z}^{\mathbf{s}}$ ) for our photorealistic data. 

Our light stage data contains paired relit images with light position and intensity from controllable LEDs. However, the ITW data lacks paired relit images and target illumination. We therefore designed a strategic mixed post-training approach, detailed in Table~\ref{tab:mixed_training}, which leverages the full flexibility of our multi-modal architecture (Sec.~\ref{sec:flexible_mm_diffusion}). Instead of one monolithic task, we train the model on different tasks tailored to the strengths of each data source.

\begin{table}[t]
\centering
\resizebox{0.99\columnwidth}{!}{%
\begin{tabular}{@{}lcccc@{}}
\toprule
Mode & Clear Latent & Noisy Latent & Global Condition & Dataset \\
\midrule
Default & - & $\mathbf{z}^{\text{all}}$ & $\mathbf{z}^{\mathbf{I}}$, $\mathbf{z}^{\mathbf{E}}$ & \texttt{Synth}, \texttt{Dome} \\
\midrule %
Rendering & $\mathbf{z}^{\mathbf{a}}$, $\mathbf{z}^{\mathbf{n}}$, $\mathbf{z}^{\mathbf{g}}$, $\mathbf{z}^{\mathbf{s}}$ & $\mathbf{z}^{\mathbf{I}_{\mathbf{E}}}$ & $\mathbf{z}^{\mathbf{E}}$ & \texttt{Synth}, \texttt{Dome} \\
Intrinsic$\rightarrow$Relit & $\mathbf{z}^{\mathbf{a}}$, $\mathbf{z}^{\mathbf{n}}$, $\mathbf{z}^{\mathbf{g}}$, $\mathbf{z}^{\mathbf{s}}$ & $\mathbf{z}^{\mathbf{I}_{\mathbf{E}}}$ & - & \texttt{ITW} \\
\midrule %
Geometry$\rightarrow$Relit & $\mathbf{z}^{\mathbf{n}}$, $\mathbf{z}^{\mathbf{g}}$, $\mathbf{z}^{\mathbf{s}}$ & $\mathbf{z}^{\mathbf{I}_{\mathbf{E}}}$, $\mathbf{z}^{\mathbf{a}}$ & $\mathbf{z}^{\mathbf{E}}$ & \texttt{Synth}, \texttt{Dome} \\ 
Relit$\rightarrow$Geometry & $\mathbf{z}^{\mathbf{I}_{\mathbf{E}}}$, $\mathbf{z}^{\mathbf{a}}$, $\mathbf{z}^{\mathbf{s}}$ & $\mathbf{z}^{\mathbf{n}}$, $\mathbf{z}^{\mathbf{g}}$ & - & \texttt{Synth} \\ 
\bottomrule
\end{tabular}
}
\vspace{-0.5em}
\caption{\textbf{Strategic training for mixing synthetic and real data.}}
\vspace{-1.5em}
\label{tab:mixed_training}
\end{table}

\begin{figure*}
    \centering
    \begin{subfigure}[b]{0.495\textwidth} %
        \centering
        \includegraphics[width=\linewidth]{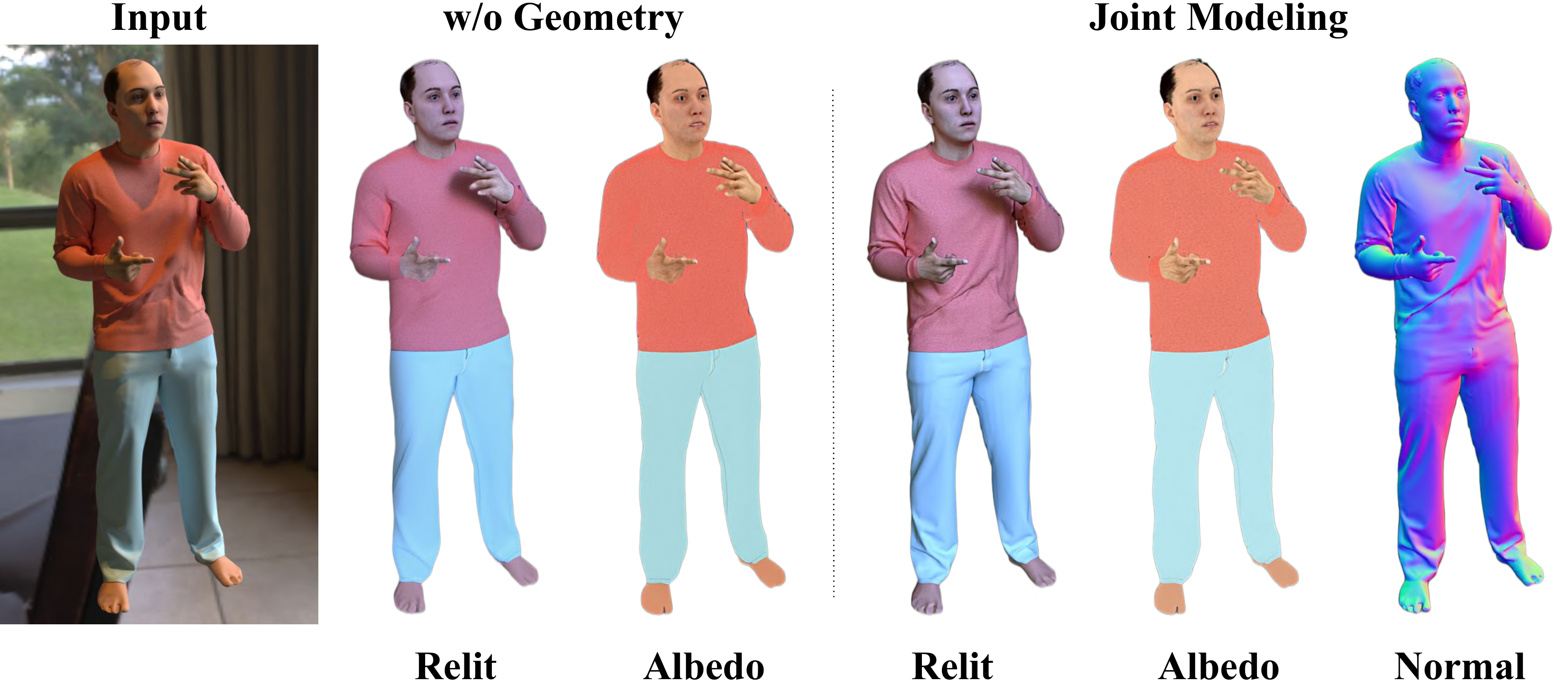}
        \caption{\textbf{Geometry is essential for relighting.} Our full "Joint Modeling" (right) captures 3D-dependent effects (wrinkles, shadows) absent in the "w/o Geometry" baseline (left).}
        \label{fig:ablation_geometry}
    \end{subfigure}
    \hfill %
    \begin{subfigure}[b]{0.495\textwidth} %
        \centering
        \includegraphics[width=\linewidth]{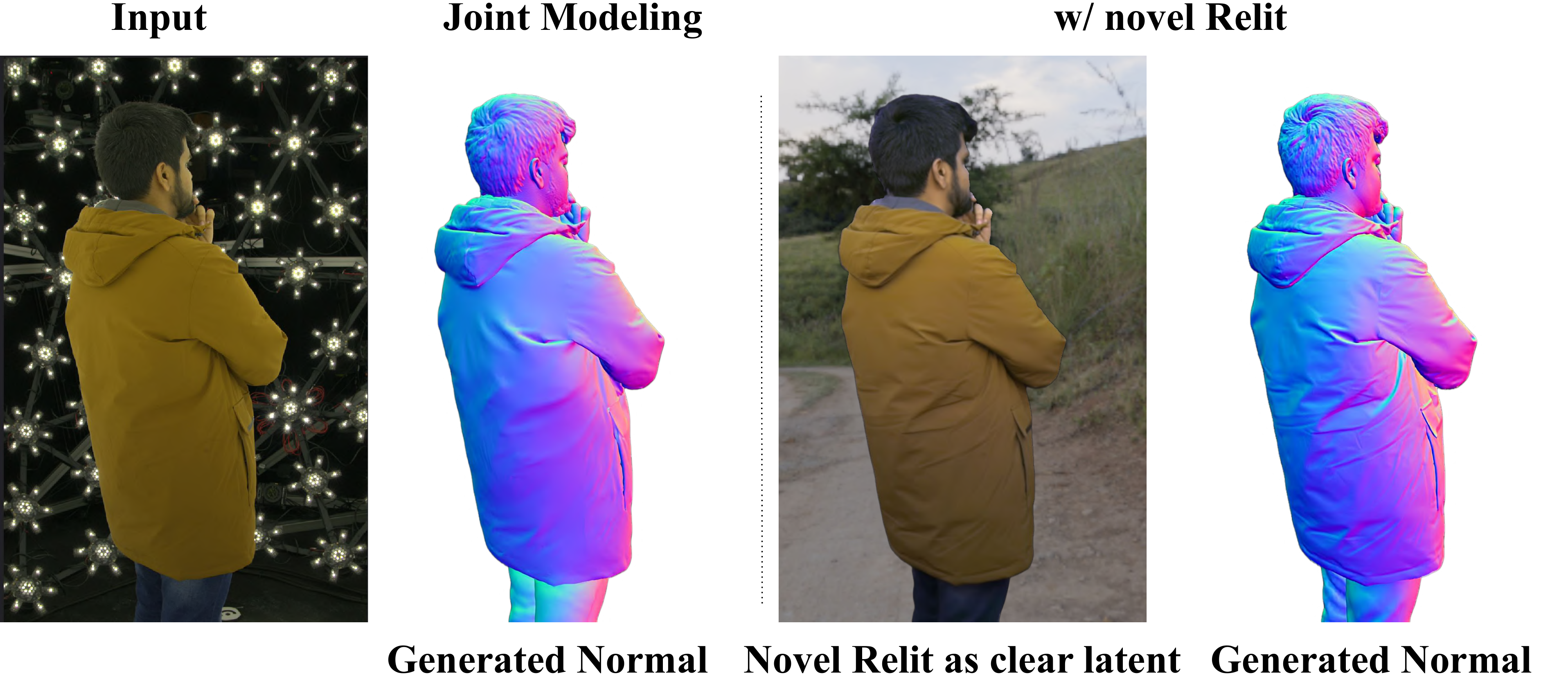}
        \caption{\textbf{Relighting provides shape-from-shading.} Given a uniformly lit input (left), a smooth normal is produced (middle). A relit as condition provides novel shading cues and enhances high-frequency normal generation (right).}
        
        \label{fig:ablation_appearance}
    \end{subfigure}

    \vspace{-1.0em} %
    
    \caption{\textbf{Ablation studies validating the synergy of joint modeling.} }
    \label{fig:ablation_synergy} %
    \vspace{-1.5em} %
\end{figure*}

On our \texttt{Synth} and \texttt{Dome} datasets, where we have complete labels, we train the full joint pipeline using the ``Default'' and ``Rendering'' modes. This teaches the model the physical relationships between geometry, intrinsics, and lighting.
For images in \texttt{ITW}, we use a specialized ``Intrinsic$\rightarrow$Relit'' task. In this mode, we provide our auto-labeled intrinsics as a clear condition (by setting $\mathbf{c}_\text{switch}{=1}$) and task the model with denoising only the original, photorealistic image $\mathbf{z}^{\mathbf{I}_{\mathbf{E}}}$. Unavailable conditions such as input $\mathbf{I}$ and illumination $\mathbf{E}$ are dropped as 0. This crucial step directly teaches the model how to synthesize photorealistic human appearance from its own intrinsic predictions, using real-world images as the target, without ever needing a ground-truth relit pair. This strategic use of our flexible architecture allows the model to learn physical accuracy from synthetic data while simultaneously learning complex, realistic appearance from in-the-wild data, all within a single unified framework.

%% file: sec/4_experiments.tex
\section{Experiments}
\label{sec:experiments}

\subsection{Experiment Settings} 

\noindent\textbf{Implementation Details}
We initialize our DiT from the inverse rendering model of DiffusionRenderer-Cosmos-7B~\cite{liang2025diffusionrenderer} and use the pretrained Cosmos causal VAE, which remains frozen.
Our training is conducted in two stages. First, we train purely on synthetic data for 30K steps to learn the fundamental disentanglement of intrinsics and relighting. 
We then use this model to "auto-label" our real-world datasets. Then, we apply our strategic mixed-data training (Sec.~\ref{sec:mixed_training}) for 10K steps, training on a combined dataset of synthetic and real data. 
A training batch contains 128 samples at resolution of $832{\times}1280$ pixels from our hybrid dataset.
The total training of two stages takes around 5 days on 64 A100 GPUs. Our 7B-parameter DiT requires 17.5 GB VRAM for inference on a single A100 GPU, generating all five modalities in approximately 35 seconds per image. Please refer to supplementary for more details in implementation and training.

\begin{figure*}
    \centering
    \includegraphics[width=\linewidth]{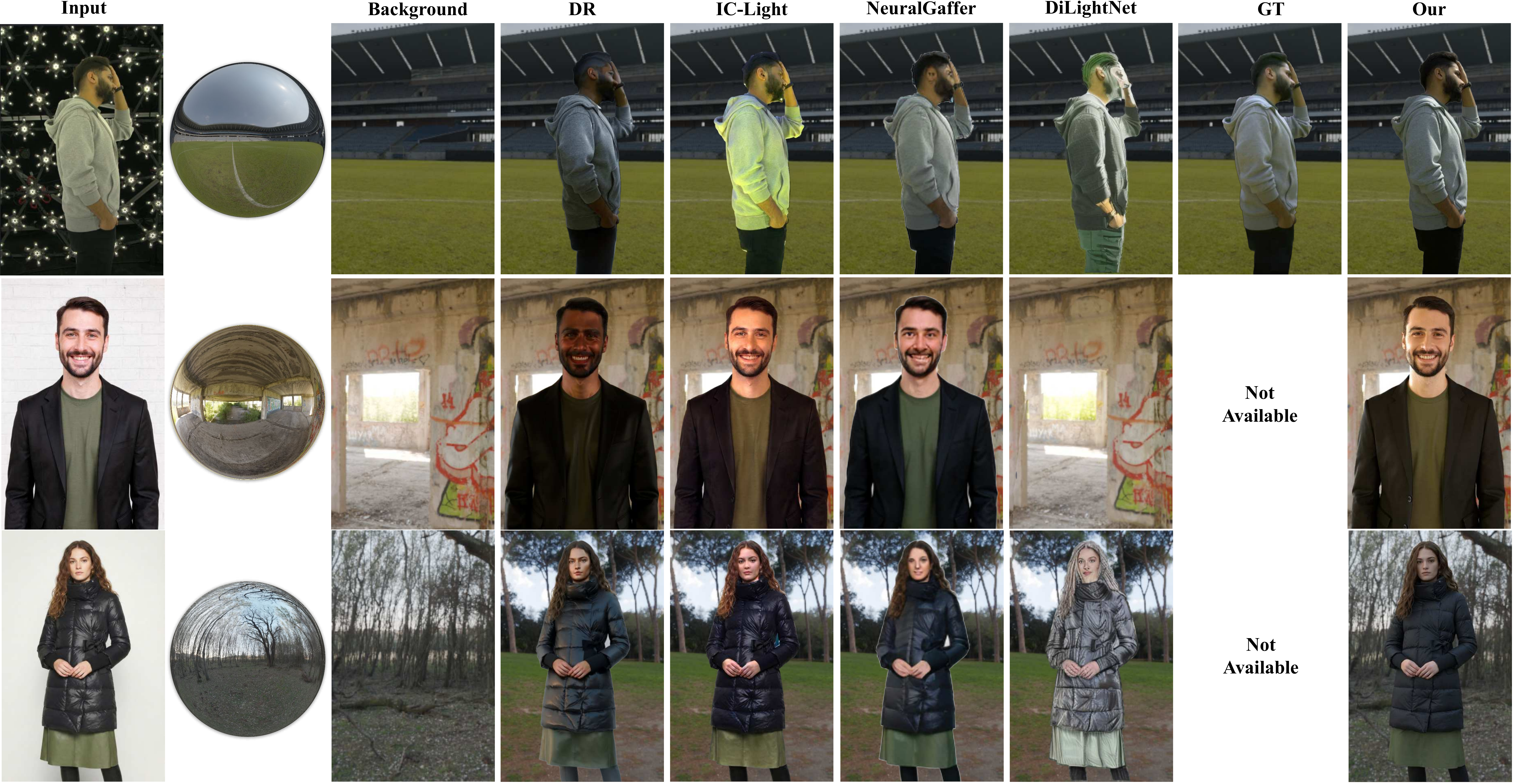}
    \vspace{-2em}
    \caption{\textbf{Qualitative comparison on relighting. }
Our model (right) produces more physically-plausible results compared to baselines on both the HumanOLAT dataset~\cite{teufelgera2025HumanOLAT} and challenging in-the-wild images. Please refer to our supplementary for more results\protect\footnotemark.
} %
    \label{fig:relit_comparison}
    \vspace{-1em}
\end{figure*}

\begin{table}[t]
\centering
  \resizebox{0.99\columnwidth}{!}{%
  \begin{tabular}{lccc|c|c}
    \toprule
& \multicolumn{3}{c}{Relighting} & \multicolumn{1}{c}{Normal} & \multicolumn{1}{c}{Point} \\ \hline
    Ablation  &  PSNR$\uparrow$ & SSIM$\uparrow$ & LPIPS $\downarrow$  & Ang. $\downarrow$  & CD. $\downarrow$ \\
\hline
    w/o Geometry    & 21.19 & 0.976 & 0.0286 &  - & -  \\
    w/ GT Geometry  &  \cellcolor{heatOrange}26.96 & \cellcolor{heatRed}\textbf{0.986} & \cellcolor{heatRed}\textbf{0.0138} &  - & -  \\
    Joint Modeling  &  \cellcolor{heatRed} \textbf{27.49} &   \cellcolor{heatOrange}0.985 &   \cellcolor{heatOrange}0.0149 &   \textbf{-} &  \textbf{-} \\
    \hline
    w/o Appearance   & - & - & - & 12.24 & 1.00  \\
    w/ GT Appearance   & - & - & - & \cellcolor{heatRed}\textbf{8.55} &  \cellcolor{heatOrange}0.66  \\
    Joint Modeling   & - & - & - &  \cellcolor{heatOrange}9.10 & \cellcolor{heatRed}\textbf{0.58}  \\
    
    \bottomrule
  \end{tabular}}
  \vspace{-0.5em}
   \caption{\textbf{Ablation Study on the Synergy of Joint Modeling.}
    We validate our core claims on the synthetic dataset. First ablation proves that geometry is essential for high-fidelity relighting. Second ablation proves that relighting provides shape-from-shading cues that improve geometry. }
    \vspace{-1.5em}
  \label{tab:ablation_study}
\end{table}

\begin{table*}
\resizebox{1.0\textwidth}{!}{
\begin{tabular}{l c c c c | c c c c | c c c c}
 \toprule
& \multicolumn{4}{c}{Synthetic Data} & \multicolumn{4}{c}{LightStage Data} & \multicolumn{4}{c}{HumanOLAT~\cite{teufelgera2025HumanOLAT}}\\
 \hline
Method & {$\text{PSNR}$ $\uparrow$} & {$\text{SSIM}$ $\uparrow$} & {LPIPS $\downarrow$}   & $\text{RMSE} \downarrow$ & {$\text{PSNR}$ $\uparrow$} & {$\text{SSIM}$ $\uparrow$} & {LPIPS $\downarrow$}   & $\text{RMSE} \downarrow$ & {$\text{PSNR}$ $\uparrow$} & {$\text{SSIM}$ $\uparrow$} & {LPIPS $\downarrow$}   & $\text{RMSE} \downarrow$ \\
 \hline
IC-Light   &  \cellcolor{heatYellow}$18.49$   & $0.880$ & \cellcolor{heatYellow}$ 0.113$ & $0.222 $ & \cellcolor{heatYellow}$20.90$ & $0.859$  &$0.119$  & \cellcolor{heatYellow}$0.172$ & \cellcolor{heatYellow}$19.79$  &\cellcolor{heatYellow}$ 0.930$ & \cellcolor{heatOrange}$0.086$ & \cellcolor{heatYellow}$ 0.185$   \\
NeuralGaffer     &  $18.84$   & \cellcolor{heatOrange}$0.929$ & \cellcolor{heatOrange}$0.105$ & \cellcolor{heatYellow}$0.220$ & $18.88$ & \cellcolor{heatOrange}$0.893$  & \cellcolor{heatOrange}$0.097$  & $0.222$ & \cellcolor{heatOrange}$20.77$  & \cellcolor{heatRed}$\textbf{0.948}$ & \cellcolor{heatRed}$\textbf{0.085}$ & \cellcolor{heatOrange}$0.165$   \\
DiLightNet    &  $14.51$   & $0.859$ & $ 0.121$ & $0.355$ & $18.03 $ & \cellcolor{heatYellow}$0.870$  & \cellcolor{heatYellow}$0.105$  & $0.243$ & $14.23$  &$0.905$ & $0.094$ & $0.337$   \\
DiffusionRenderer   &  \cellcolor{heatOrange}$19.28$   & \cellcolor{heatYellow}$0.886$ & $0.119$ & \cellcolor{heatOrange}$0.203$ & \cellcolor{heatOrange}$21.09$ & $0.862$  &$0.120$  & \cellcolor{heatOrange}$0.169$ & $17.58$  &$0.899$ & $0.100$ & $0.229$   \\
\hline
 \textbf{\textit{GeoRelight}}   &  \cellcolor{heatRed}$\textbf{27.22}$   & \cellcolor{heatRed}$\textbf{0.941}$ & \cellcolor{heatRed}$\textbf{0.057}$ & \cellcolor{heatRed}$\textbf{0.079}$ & \cellcolor{heatRed}$\textbf{25.87}$ & \cellcolor{heatRed}$\textbf{0.907}$  & \cellcolor{heatRed}$\textbf{0.086}$  & \cellcolor{heatRed}$\textbf{0.092}$ & \cellcolor{heatRed}$\textbf{21.17}$  &\cellcolor{heatOrange}${0.935}$ &\cellcolor{heatOrange}${0.086}$ & \cellcolor{heatRed}$\textbf{0.161}$ \\
\bottomrule
\end{tabular}
}
\centering
\vspace{-0.5em}
\caption{\textbf{Relighting evaluation for human images.} Our method achieves state-of-the-art relighting performance. Although NeuralGaffer achieves higher metric on HumanOLAT, it cannot faithfully recover details in relit images. More comparisons are in supplementary.}
\label{tab:relighting_comparison}
\vspace{-2em}
\end{table*}

\begin{figure*}
    \centering
    \includegraphics[width=\linewidth]{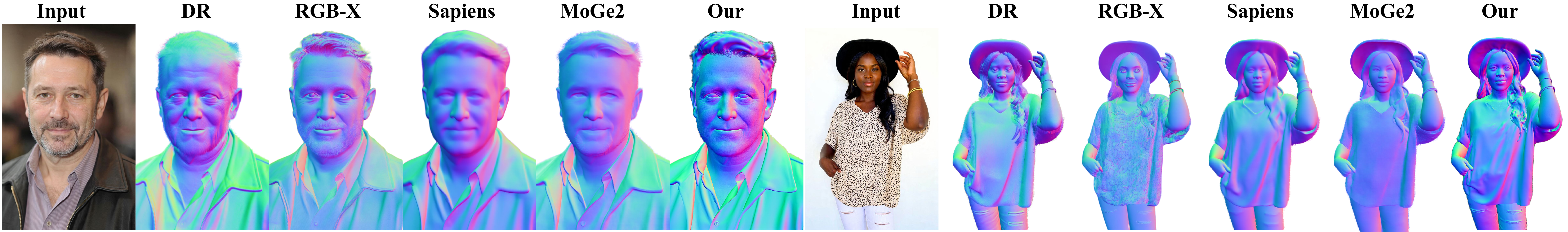}
    \vspace{-2em}
    \caption{\textbf{Qualitative comparison of estimated normal.} Our model outperforms all baselines and consistently achieves sharper and high-frequency details such as eyes, skin, and hair. Please zoom in for details.}
    \vspace{-1em}
    \label{fig:normal_comparison}
\end{figure*}

\noindent\textbf{Evaluation Details}
First we evaluate on synthetic data, which contains 42 identities with ground-truth for all modalities from our synthetic dataset. Then, we evaluate on 50 identities from our light stage captures. We also evaluate on HumanOLAT~\cite{teufelgera2025HumanOLAT}. Following official instruction, we generate ground-truth relit images by combining provided OLAT data. We provide extensive qualitative results on in-the-wild images here and in our supplementary.

For relighting, we quantitatively compare against open-source baselines~\cite{liang2025diffusionrenderer, jin2024neuralgaffer, zeng2024dilightnet, zhang2025iclight} with PSNR, LPIPS, and RMSE computed on foreground subject. We also provide qualitative comparison and user study with~\cite{mei2025luxpostfacto}. 
For geometry, we compare against feed-forward point map predictors~\cite{wang2025vggt, wang2025moge2}. We report Chamfer distance and F-score on the normalized and ICP-aligned~\cite{arun1987icp} geometry to ground-truth shape.
For intrinsics, we evaluate normal estimation against~\cite{khirodkar2024sapiens, wang2025moge2, liang2025diffusionrenderer, zeng2024rgbx} with angular error and RMSE. For albedo estimation we compare against~\cite{liang2025diffusionrenderer, zeng2024rgbx} using PSNR, SSIM, and LPIPS. Please refer to our supplementary for detailed evaluation protocol.

\subsection{Ablation Studies: Validating our Design}

\begin{figure*}
    \centering
    \includegraphics[width=\linewidth]{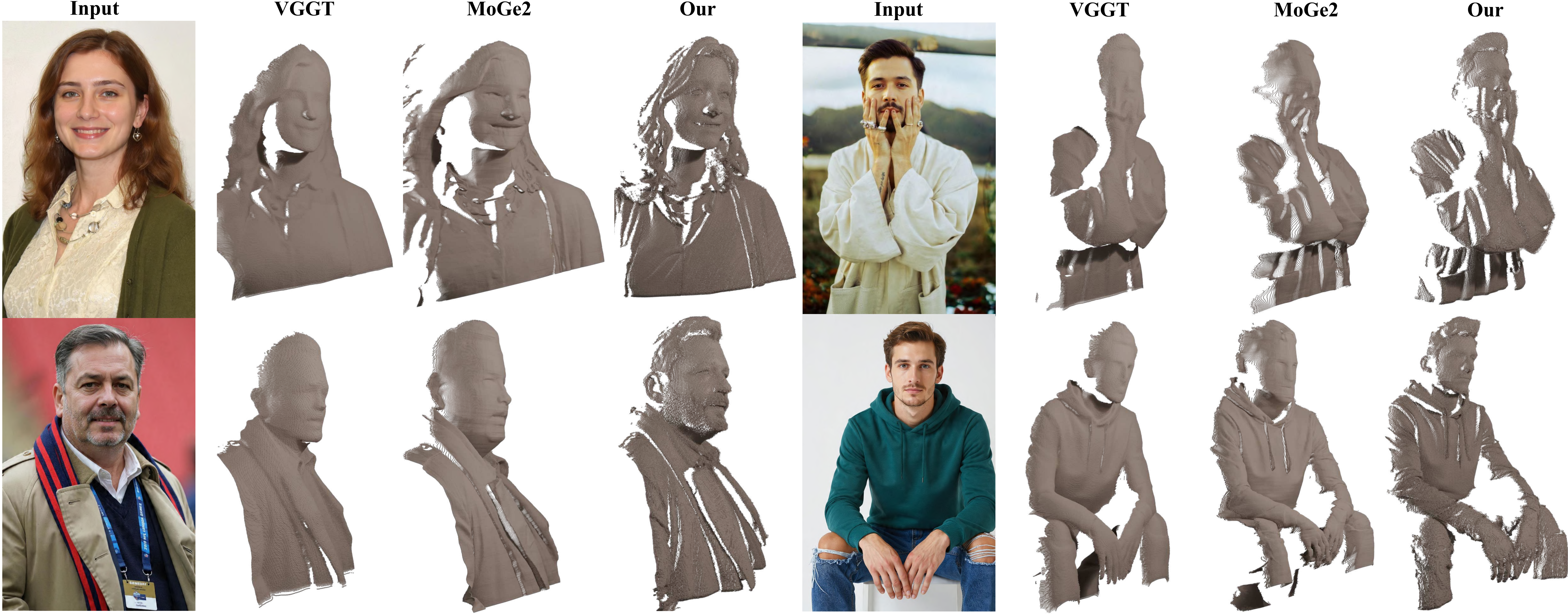}
    \vspace{-2em}
    \caption{ \textbf{Qualitative comparison on geometry reconstruction.}
Our joint model (right) reconstructs fine-grained 3D shapes. In contrast, specialized geometry estimators like VGGT~\cite{wang2025vggt} and MoGe2~\cite{wang2025moge2} produce distorted or over-smoothed point clouds on these in-the-wild images, demonstrating the superior performance of high-frequency details modeling of our iNOD with latent generative models.}
    \label{fig:geometry_comparison}
    \vspace{-1.5em}
\end{figure*}

\begin{figure}
    \centering
    \setlength{\unitlength}{\linewidth} 
    
    \begin{picture}(1, 0.4) %
        \put(0,0){\includegraphics[width=\linewidth]{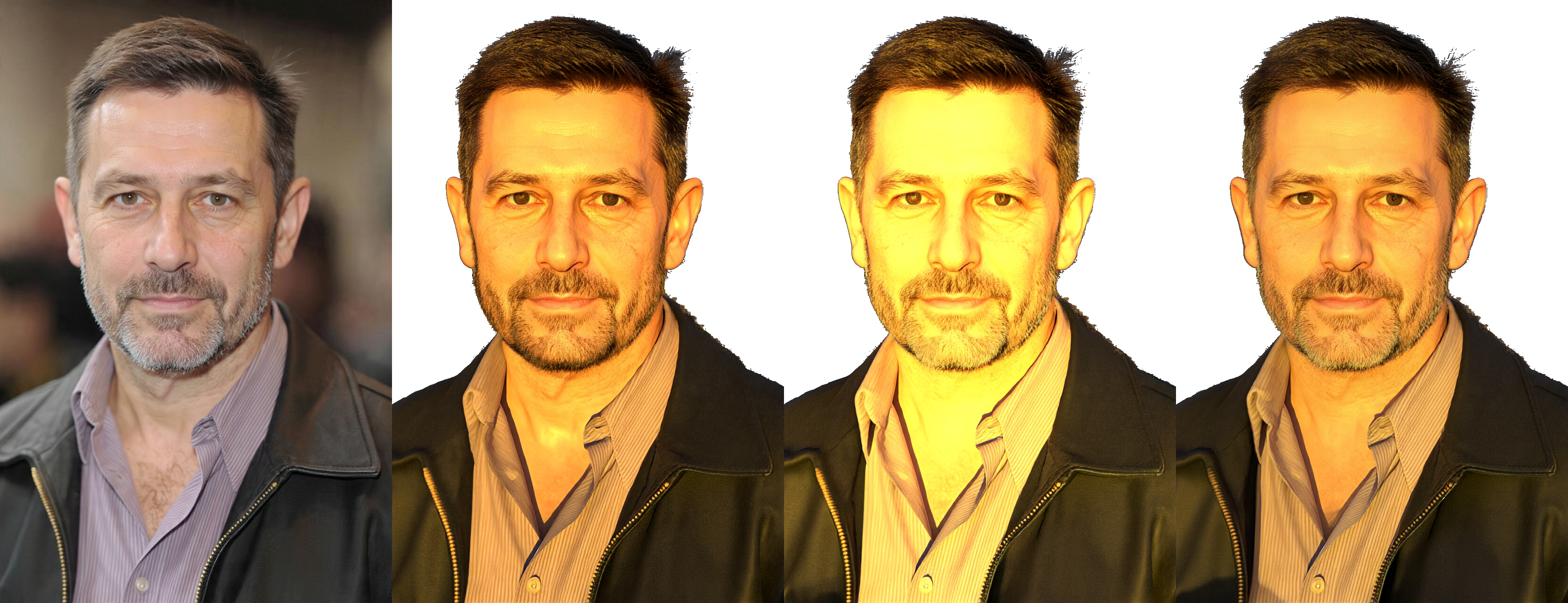}}
        
        \put(0.17, 0.02){\color{white}\scriptsize \textbf{Input}}
        \put(0.42, 0.02){\color{white}\scriptsize \textbf{Synth}}
        \put(0.67, 0.02){\color{white}\scriptsize \textbf{\shortstack{Synth \\ Dome}}}
        \put(0.92, 0.02){\color{white}\scriptsize \textbf{\shortstack{Synth \\ Dome \\ ITW}}}
    \end{picture}

    \vspace{-.5em} %
    \caption{\textbf{Benefit of In-the-Wild Data.} Using only \texttt{Synth} uncovers gaps in the data like the lack of mixed colored beards. Adding \texttt{Dome} data fixes that but produces unrealistic brightness (middle) due to the unnatural LED activation (either very sparse or fully lit) in light stage captures. Adding large-scale \texttt{ITW} data corrects this bias, yielding balanced and realistic lighting (right).}
    \label{fig:itw_benefit}
    \vspace{-1.em}
\end{figure}

Before comparing to external baselines, we first validate the core claims of our paper: that our multi-modal synergetic generation of geometry and relighting, and mixed-data strategy are helpful for high-quality performance.

\noindent\textbf{Geometry is essential for Relighting} We compare three different modes: Joint Modeling (generating geometry and relighting together); w/GT Geometry (take ground-truth geometry as clear latent and set up switch $\mathbf{c}_{\text{switch}} = 1$ to generate relit image); w/o Geometry (only generate relit without joint generation of geometry). Tab.~\ref{tab:ablation_study} and Fig.~\ref{fig:ablation_geometry} clearly demonstrates how important the geometry information is for relighting: joint modeling of geometry leads to fine-grained details such as shadow and clothing wrinkles, while the "w/o Geometry" mode fails to capture these geometry-dependent effects. Please refer to supplementary for details.

\noindent\textbf{Relighting provides shape-from-shading} To prove this, we conduct the reverse experiment. We compare our "Joint Modeling" mode against generation "w/o Appearance" and "w/ GT Appearance". Figure~\ref{fig:ablation_appearance} shows that when the model is given a ground-truth relit image from HumanOLAT as a clear latent condition, it generates a much more detailed and accurate normal map. This demonstrates that the appearance of shading provides critical information that our model successfully uses to refine geometry estimation.

\noindent\textbf{In-the-wild data corrects lighting bias} We further analyze the specific contribution of our in-the-wild (\texttt{ITW}) data. As shown in Fig.~\ref{fig:itw_benefit}, training with only \texttt{Synth}+\texttt{Dome} data already improves realism over synthetic-only training, but introduces a ``dark bias'' in the relit outputs. This is because the light stage captures use sparse LED activations (10-20 out of 1024 lights), resulting in predominantly dark training images that skew the model's output distribution. Incorporating large-scale \texttt{ITW} data with diverse, real-world lighting conditions corrects this bias, producing relit images with balanced and natural illumination.

\noindent\textbf{Generalizing iNOD Beyond Humans} While we focus on human subjects in this work, iNOD is a general-purpose representation that is not inherently limited to any specific category. In supplementary, we validate that iNOD faithfully encodes and decodes 3D geometry for subjects with increasing Depth-to-Height (D:H) ratios, from humans (${\sim}$0.1) to objects like buildings (3.0). The representation remains effective as long as the D:H ratio is moderate; extremely elongated scenes (e.g., tunnels with D:H ${>}$10) would compress the depth range, potentially degrading reconstruction quality. Consequently, the generalizability of iNOD unlocks the potential of our framework for joint relighting, intrinsic decomposition, and 3D reconstruction in broader object-centric domains.

\subsection{Comparison with State-of-the-art}

\footnotetext{The images in Figures \ref{fig:teaser}, \ref{fig:relit_comparison} (second and third row),  \ref{fig:normal_comparison}, \ref{fig:geometry_comparison}, and \ref{fig:itw_benefit} are synthetic images created using a generative model.}

\noindent\textbf{Relighting Performance}
We first evaluate our primary task of relighting. Tab.~\ref{tab:relighting_comparison} shows the quantitative comparison on our synthetic, light stage, and HumanOLAT datasets. Our method significantly outperforms all open-source general-purpose baselines, including DiffusionRenderer (DR), NeuralGaffer, and DiLightNet, across all datasets and metrics. 
We observe that the gap on HumanOLAT is smaller than our hold-out evaluation data, which can be possible due to the artifacts in the ground-truth. 
Due to data license limitation, we could not obtain quantitative results for closed-source methods~\cite{mei2025luxpostfacto}, we conduct a user study on in-the-wild inference in supplementary. Fig.~\ref{fig:relit_comparison} provides a qualitative comparison, showing that our model produces more realistic and detailed results.
Our qualitative results on in-the-wild images (Fig.~\ref{fig:relit_comparison}, row 3) confirm this. Our method generates physically-plausible shading and shadows that are consistent with the new illumination, while other methods like IC-Light and DiLightNet produce artifacts or fail to match the illumination.

\noindent\textbf{Geometric \& Intrinsic Estimation Performance}
We report the quantitative comparison of fullbody and portrait reconstruction on our synthetic data in Tab.~\ref{tab:point_comparison}. We achieve superior performance compared to SoTA general purpose reconstructor VGGT~\cite{wang2025vggt} and MoGe2~\cite{wang2025moge2}. Fig.~\ref{fig:geometry_comparison} qualitatively confirms this, showing our model captures detailed shapes where others produce distorted or over-flattend results.
Similarly, Tab.~\ref{tab:albedo_comparison} demonstrates our leadership in estimation of albedo and surface normal. Fig.~\ref{fig:normal_comparison} shows that our model can recover fine-grained surface details in clothing, skin, and hair.

\begin{table}[t]
\centering
  \resizebox{0.99\columnwidth}{!}{%
  \begin{tabular}{@{}lcccc cc c@{}}
    \toprule
    Method  &  Acc.$\downarrow$ & Comp.$\downarrow$ & CD $\downarrow$  & F-Score $\uparrow$  &  Prec. $\uparrow$ &$\text{Rec.}$ $\uparrow$ \\
\midrule
    VGGT              & \cellcolor{heatOrange} 4.06 & \cellcolor{heatOrange}2.68 & \cellcolor{heatOrange} 3.37 & 21.05 & \ 18.87 & 23.98 \\
    MoGe2                  & 4.07 & 3.02 & 3.54 &  \cellcolor{heatOrange}23.96 &  \cellcolor{heatOrange}23.71 &  \cellcolor{heatOrange}24.41 \\
    \hline
    \textbf{\textit{GeoRelight}} & \cellcolor{heatRed}\textbf{0.71} & \cellcolor{heatRed}\textbf{0.82} & \cellcolor{heatRed}\textbf{0.766} & \cellcolor{heatRed}\textbf{81.56} & \cellcolor{heatRed} \textbf{84.24} &  \cellcolor{heatRed} \textbf{79.09} \\ 
    \bottomrule
  \end{tabular}}
 \vspace{-0.5em}
   \caption{\textbf{Geometric Shape Estimation evaluation}.
\textbf{GeoRelight} outperforms specialized, single-task SoTA geometry estimators.}
  \label{tab:point_comparison}
  \vspace{-1.em}
\end{table}

\begin{table}[t]
\centering
  \resizebox{0.99\columnwidth}{!}{%
  \begin{tabular}{lccc|cc}
    \toprule
& \multicolumn{3}{c}{Albedo} & \multicolumn{2}{c}{Normal} \\ \hline
    Method  &  PSNR$\uparrow$ & SSIM$\uparrow$ & LPIPS $\downarrow$  & Ang. $\downarrow$  & RMSE $\downarrow$ \\
\hline
    Sapiens                  & - & - & - &  \cellcolor{heatOrange}13.99 & \cellcolor{heatOrange}0.319  \\
    MoGe2                  & - & - & - & \cellcolor{heatYellow}16.33 & \cellcolor{heatYellow}0.344  \\
    \hline
    RGB-X              & \cellcolor{heatYellow}15.45 & \cellcolor{heatOrange}0.909 & \cellcolor{heatOrange}0.093 & 21.12 & 0.420  \\
    DiffusionRenderer                  & \cellcolor{heatOrange}19.46 & \cellcolor{heatYellow}0.892 & \cellcolor{heatYellow}0.115 & 20.44 &  0.420  \\
    \hline
    \textbf{\textit{GeoRelight}}   &  \cellcolor{heatRed} \textbf{28.07} &  \cellcolor{heatRed} \textbf{0.943} &  \cellcolor{heatRed} \textbf{0.057} &  \cellcolor{heatRed} \textbf{8.64} & \cellcolor{heatRed} \textbf{0.211} \\
    \bottomrule
  \end{tabular}}
  \vspace{-0.5em}
   \caption{\textbf{Albedo and Normal disentanglement evaluation.}
   \textbf{GeoRelight} achieves state-of-the-art albedo and normal estimation.}
  \label{tab:albedo_comparison}
  \vspace{-1.5em}
\end{table}

%% file: sec/5_summary.tex
\section{Conclusion}
\label{sec:summary}

In this paper, we propose ~\textbf{GeoRelight}, a unified framework that, for the first time, jointly generates photorealistic relit images and reconstructs high-fidelity 3D geometry within a single multi-modal diffusion transformer.
Our solution is enabled by three key contributions: (1) a flexible DiT architecture that processes modalities in parallel; (2) iNOD, a novel, distortion-free geometry representation compatible with pretrained VAEs; and (3) a strategic mixed-data training method that leverages Dome and in-the-wild data to bridge the synthetic-real gap. Our experiments proved our core thesis: joint modeling is synergistic, and our model achieves state-of-the-art results in not just relighting, but also in geometry and intrinsic estimation, outperforming specialized, single-task methods.
While GeoRelight is a robust single-image method, it does not yet model temporal consistency. As our core DiT is built from a video model, a clear direction for future work is to extend our joint framework to video relighting and reconstruction.
Besides, joint estimating geometry and appearance for relightable 3D human generation~\cite{xue2024human3diffusion, xue2025infinihuman, xue2025gen3diffusion} can be a downstream application.

%% file: sec/supp_body.tex
\DeclareRobustCommand{\code}[1]{\texttt{\detokenize{#1}}}

\section*{Supplementary Material}
In this supplementary document, we provide further details about the implementation details, training data, evaluation protocol, etc. We also show additional comparison with state-of-the-art approaches. Please refer to our supplementary video for additional visualization of our results. Finally, we discuss limitations of our model and potential future works.

\section{Additional Implementation Details}

\begin{figure}[b]
    \centering
    \includegraphics[width=1.0\linewidth]{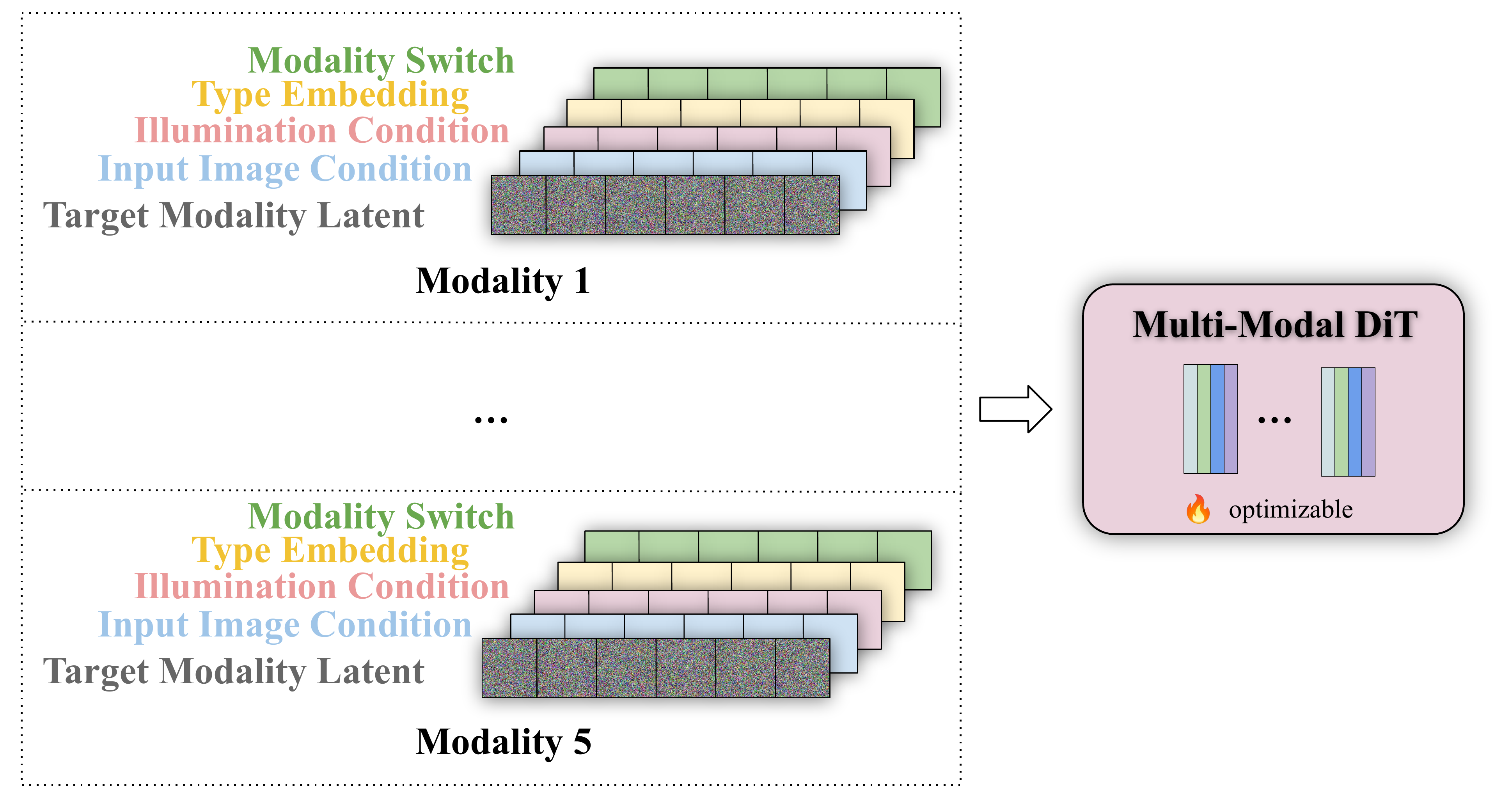}
    \caption{ \textbf{Conditioning on the modality latent.}  Each modality latent after conditioning have the shape $\mathbb{R}^{H\times W\times C(16+16+3*16+3+1)}$. Different modalities are concatenated "temporal-wise" to a sample $\mathbb{R}^{M\times H\times W\times C}$ in one batch.}
    \label{fig:supp_conditioning}
\end{figure}
This section provides further details on our network architecture and training procedure, expanding on Sec. 4.1 of the main paper.
\subsection{Model Architecture}
\paragraph{Base Model} Our model's architecture is a Diffusion Transformer (DiT)~\cite{Peebles2022DiT}. We initialize our weights from the publicly available inverse rendering model of DiffusionRenderer-Cosmos-7B~\cite{liang2025diffusionrenderer}, which is built upon the Cosmos-Predict-1 7B DiT~\cite{nvidia2025cosmospredict1}.

\paragraph{Causal VAE Latent Space} We operate in the latent space of the pretrained Cosmos causal VAE~\cite{nvidia2025cosmospredict1}, which can process both image or videos. The VAE remains frozen during our training. This VAE has a spatial compression factor of 8. Therefore, our input images at a training resolution of $832 \times 1280$ pixels are encoded into a latent space of $104 \times 160$. The channel dimension is 16. All five target modalities ($\mathbf{z}^\mathbf{a}, \mathbf{z}^\mathbf{n}, \mathbf{z}^\mathbf{g}, \mathbf{z}^\mathbf{s}, \mathbf{z}^{\mathbf{I}_\mathbf{E}}$) and the global image condition ($\mathbf{z}^\mathbf{I}$) and illumination condition ($\mathbf{z}^\mathbf{E}$) share this latent resolution.

\subsection{Detailed DiT Conditioning} As illustrated in Fig.~\ref{fig:supp_conditioning}, our architecture adapts a standard DiT block to accept our multi-modal conditioning. The total input to the DiT for a single modality $m$ is a channel-wise concatenation of several features that are added to the noisy latent $z^m_{\tau}$ before the first DiT block.

\begin{itemize}
    \item Global Condition $\mathbf{z}^{\mathbf{I}}$: The VAE-encoded latent of the input image $\mathbf{I}$. This is concatenated to all five target modalities to provide shared identity and shape information.
    \item Illumination Condition ($\mathbf{z}^{\mathbf{E}}$): The VAE-encoded latent of the three LDR light maps ($\mathbf{E}_{ldr}, \mathbf{E}_{log}, \mathbf{E}_{dir}$). This is concatenated only to the relit image latent $\mathbf{z}^{\mathbf{I}_\mathbf{E}}$ and is zero-padded for all other modalities.
    \item Modality Type Embedding ($\mathbf{c}_{\text{modal}}$): A learnable embedding of shape $\mathbb{R}^{M \times C_{\text{type}}}$ (where $M$=5, $C_{\text{type}}=3$) that is broadcast to the latent spatial resolution and concatenated. This informs the model which modality it is currently processing (e.g., albedo vs. normal).
    \item Modality Switch Mask ($\mathbf{c}_{\text{switch}}$): A binary mask of shape $\mathbb{R}^{H \times W \times 1}$ that is broadcast and concatenated. It signals whether the modality is a "clear" condition (1) or a "noisy" target (0).
\end{itemize}
These concatenated features $\in \mathbb{R}^{H\times W\times (16+16+3*16+3+1)}$ are then processed by the DiT. We replace the original 3D RoPE positional embeddings with a shared 2D RoPE~\cite{su2023rope} that is applied to all modalities. This ensures that spatial position (x,y) is aligned across all modalities, which is critical for learning cross-modal correlations.

\subsection{Training Details}

\paragraph{Hardware and Optimizer} Our model is trained on 64 NVIDIA A100 (80GB) GPUs. We use the AdamW~\cite{kingma2017adam} optimizer with a constant learning rate of $2 \times 10^{-5}$, a weight decay of 0.01, and enable bfloat16 mixed-precision training. Our total effective batch size is 128, distributed across all GPUs.

\paragraph{Two Stage Training} As described in the main paper (Sec 4.1), our training is conducted in two distinct stages:

\begin{enumerate}
    \item Stage 1 (Synthetic Pre-training): We first train the model for 30,000 steps exclusively on our synthetic \texttt{Synth} dataset. This stage teaches the model the fundamental disentanglement of intrinsics, geometry, and lighting. The resulting model from this stage is then used as our "auto-labeler" for the real-world datasets.

    \item Stage 2 (Mixed Post-Training): We then finetune the model from Stage 1 for an additional 10,000 steps. In this stage, we train on a combined hybrid dataset of synthetic, light stage, and in-the-wild data, applying the strategic tasks outlined in our main paper's Table 1. The batch is composed by sampling from our datasets with the following probabilities: 33\% \texttt{Synth}, 35\% \texttt{Dome}, 32\% \texttt{ITW}.
\end{enumerate}

\subsection{Inference Runtime}
We report our method's inference speed on a single NVIDIA A100 80GB GPU.
For a single input image at our test resolution of $832 \times 1280$, the full joint-generation process takes approximately 35 seconds. This time include 35 diffusion sampling steps, VAE decoding time, and I/O time for saving the results.

\subsection{Ablation Details}
We provide more details about implementation of ablation study in main paper section 4.2. The quantitative ablation study was reported on full-body subset of our synthetic evaluation dataset because full-body images contain more wrinkles compared to face-only subset.

\paragraph{Geometry is essential for Relighting} We compare three distinct modes to validate this claim:
\begin{itemize}
    \item Joint Modeling: The model generates all five modalities (relit image, albedo, normal, segmentation, and geometry) simultaneously from noise.
    \item w/ GT Geometry: We replace the initial noise of the geometry modalities (iNOD, normal, segmentation) with the ground-truth latent representations. We set their corresponding switch masks to $c_\text{switch}=1$ (condition) and keep them fixed (clear) during the reverse sampling process, forcing the model to generate the relit image conditioned on perfect geometry.
    \item w/o Geometry: We disable the generation of geometry modalities. To keep the input tensor dimensions consistent with our architecture, we replace the geometry latents with zero-tensors. This formulation is analogous to the dropped-condition scenario in classifier-free guidance~\cite{ho2022classifierfreediffusionguidance}, effectively removing geometric guidance from the generation process.
\end{itemize}

\paragraph{Relighting helps Shape-from-Shading.} To validate the synergistic effect of appearance on geometry, we compare the following modes:
\begin{itemize}
    \item \textbf{w/o Appearance:} We isolate the geometry generation by suppressing the relighting modality. Similar to the ``w/o Geometry'' setting, we replace the relit image latent $z^{I_E}$ and the illumination condition $z^E$ with zero-tensors, effectively forcing the model to hallucinate geometry from the input image $I$ alone, without any shading cues from the new lighting context.

    \item \textbf{w/ GT Appearance:} We provide the ground-truth relit image as a strong conditioning signal. We encode the GT relit image into latent space and set its switch mask to $c_{\text{switch}}=1$. This allows the model to explicitly use the shading and shadows present in the target relit image to refine the surface normals and iNOD geometry.
\end{itemize}

\section{Detailed Data Creation and Sources}
In this section, we provide a detailed breakdown of the three data sources used in our hybrid dataset, as first introduced in Sec. 3.4 of the main paper and visually summarized in our main paper's Figure 4.

\subsection{Synthetic Data}
The synthetic dataset forms the foundation of our method. Its primary purpose is to provide a large corpus of physically-accurate, fully-labeled data to train our Stage 1 model. We create 1000 high-quality, artist-created 3D human meshes, which contain detailed facial components (eyes, teeth, hair). We augment it to 8000 identities by rigging for pose variation.  We pair these with a library of high-quality PBR texture maps for clothing (albedo, normal, roughness).

Our system is built with Blender, and images are rendered with the Cycles renderer. For each of the 8000 appearance samples, we render it under 400 random HDR environment maps, creating a large-scale training set. For each rendered scene, we save all five of our target modalities: the relit image ($\mathbf{I}_{\mathbf{E}}$), albedo ($\mathbf{a}$), segmentation mask ($\mathbf{s}$), surface normals ($\mathbf{n}$), and our novel iNOD geometry ($\mathbf{g}$), which is computed from the ground-truth metric depth and camera intrinsics.

As noted in our main paper, this synthetic-only model yields satisfactory intrinsics but lacks photorealism in the final relit image, which motivates our hybrid-data approach.

\subsection{Light Stage Data}
The Dome dataset is our high-quality, real-world dataset that acts as a bridge between synthetic data and in-the-wild images. It is sourced from the Codec Avatar Studio~\cite{martinez2024castudio} and Relightable Full-Body Gaussian Codec Avatars~\cite{wang2025fullbodyavatar} datasets.

\begin{figure}
    \centering
    \includegraphics[width=1.0\linewidth]{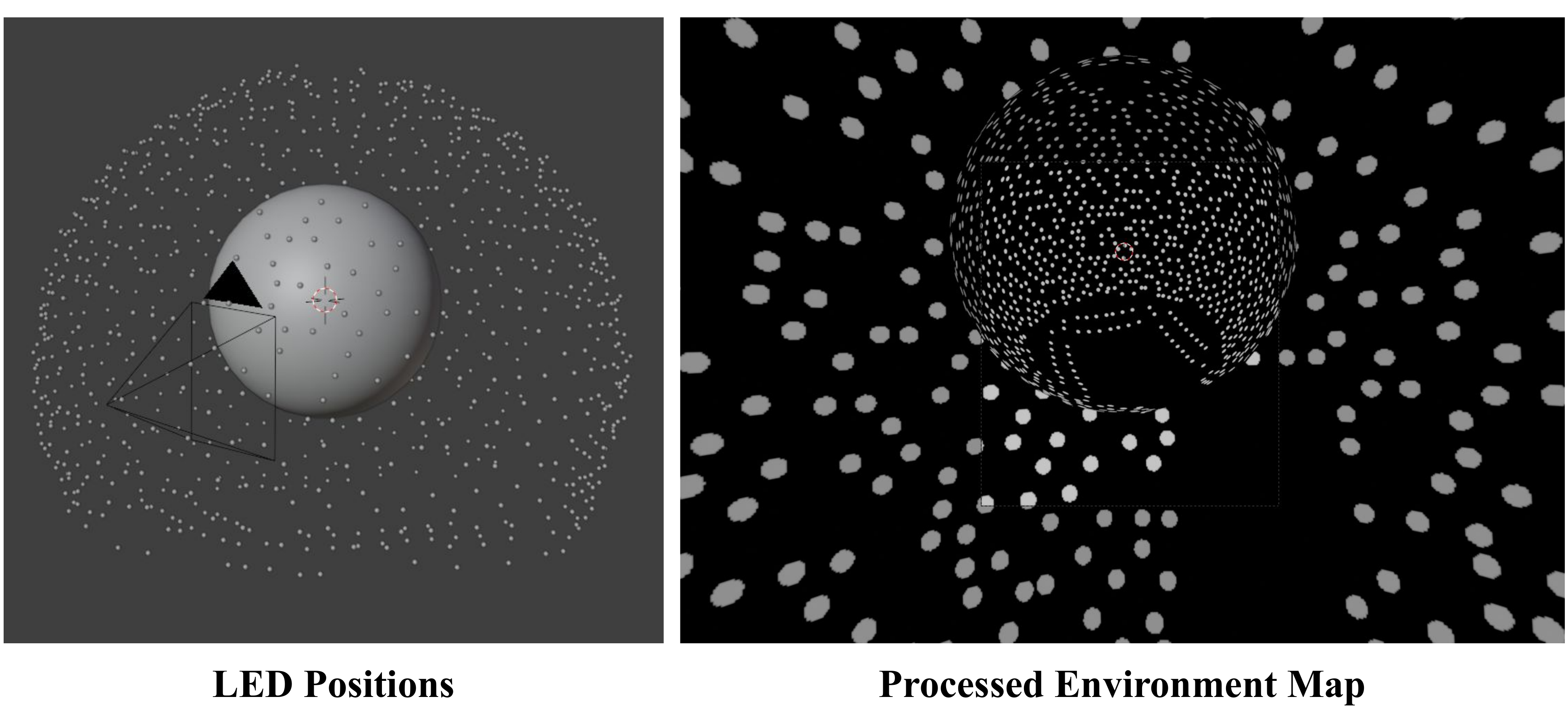}
    \caption{ \textbf{Processed Environment Illumination from LightStage.} From the 3-dimensional LED positions, we project it to a latlong image to model the environement map.}
    \label{fig:supp_lightstage}
\end{figure}

\paragraph{Paired Data Creation} This dataset is critical because its light stage capture setup, which employs 1024 individually controllable white LED light sources with known locations, allows us to create paired real-world training data. The captured videos consist of alternating fully-lit frames and partially-lit frames (random 10-20 lights) at 60 Hz.

Due to fast motion, we cannot reliably warp frames to perform traditional image-based relighting as in~\cite{pandey2021totalrelighting, kim2024switchlight, mei2025luxpostfacto}. Hence, we treat neighboring frames as the paired input and relit image. To generate the corresponding illumination condition $\mathbf{E}$, we map the 3D LED light locations and intensities from the partially-lit frame to an equirectangular image (Fig.~\ref{fig:supp_lightstage}, simulating a sparse environment map. This process provides us with realistic, photometrically-aligned pairs of (Input, Target Light, Target Relit Image), which we use for the "Default" and "Rendering" training modes (Table 1 in main paper).

\subsection{In-the-wild Data}
To further enhance diversity and photorealism, we use a large-scale in-the-wild dataset. We sample 10K human images from the CosmicMan~\cite{li2025cosmicman} and ReLaion~\cite{laion2024relaion} datasets as our data pool.

\paragraph{Label Scarcity} This dataset is defined by what it lacks. Unlike \texttt{Synth} and \texttt{Dome}, \texttt{ITW} images have no ground-truth intrinsics, no paired relit image, and no known illumination condition $\mathbf{E}$. It is used exclusively for our "Intrinsic$\rightarrow$Relit" task.

\subsection{Auto-Labeling Process}

\begin{figure}
    \centering
    \includegraphics[width=\linewidth]{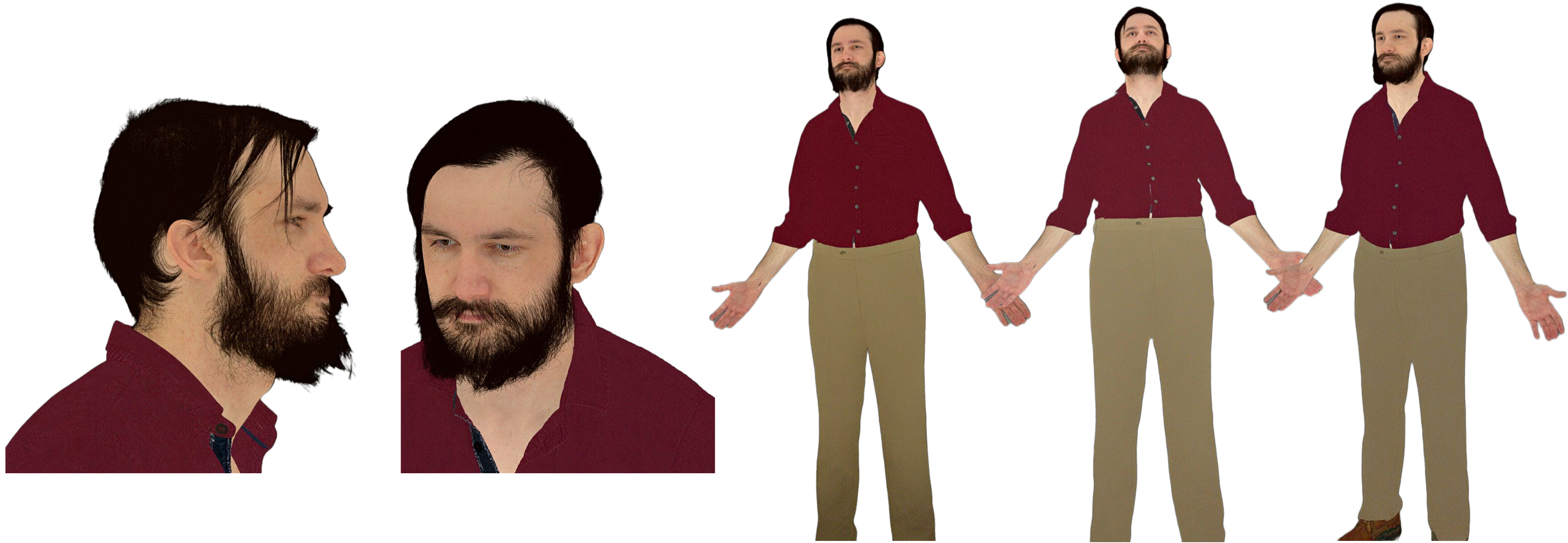}
    \caption{\textbf{Robustness of our Auto-Labeler.}
Our auto-labeled albedo (shown) and other intrinsics are consistent across multiple views of the same subject from our Dome dataset, demonstrating the high quality of our pseudo-ground-truth.}
    \label{fig:supp_albedo_consistency}
\end{figure}

The "auto-labeling" process is the important component of our strategic mixed-data training, inspired by~\cite{liang2025diffusionrenderer, mei2025luxpostfacto}. We use the model trained exclusively on our \texttt{Synth} dataset (Stage 1) as our "auto-labeler."

We run this model in inference mode over our entire Dome and ITW datasets. For each real-world image, we generate and save the four corresponding pseudo-ground-truth intrinsic maps: albedo ($\mathbf{z}^\mathbf{a}$), segmentation mask ($\mathbf{z}^\mathbf{s}$), surface normals ($\mathbf{z}^\mathbf{n}$), and our iNOD geometry ($\mathbf{z}^\mathbf{g}$). As shown in Fig~\ref{fig:supp_albedo_consistency}, this auto-labeler is surprisingly robust, generating high-quality, view-consistent intrinsics even for real-world subjects. This process is what enables us to use the Dome and ITW datasets for our advanced training tasks in Stage 2.

\section{Additional Details on iNOD}

\begin{figure}[t]
    \centering
    \includegraphics[width=1.0\linewidth]{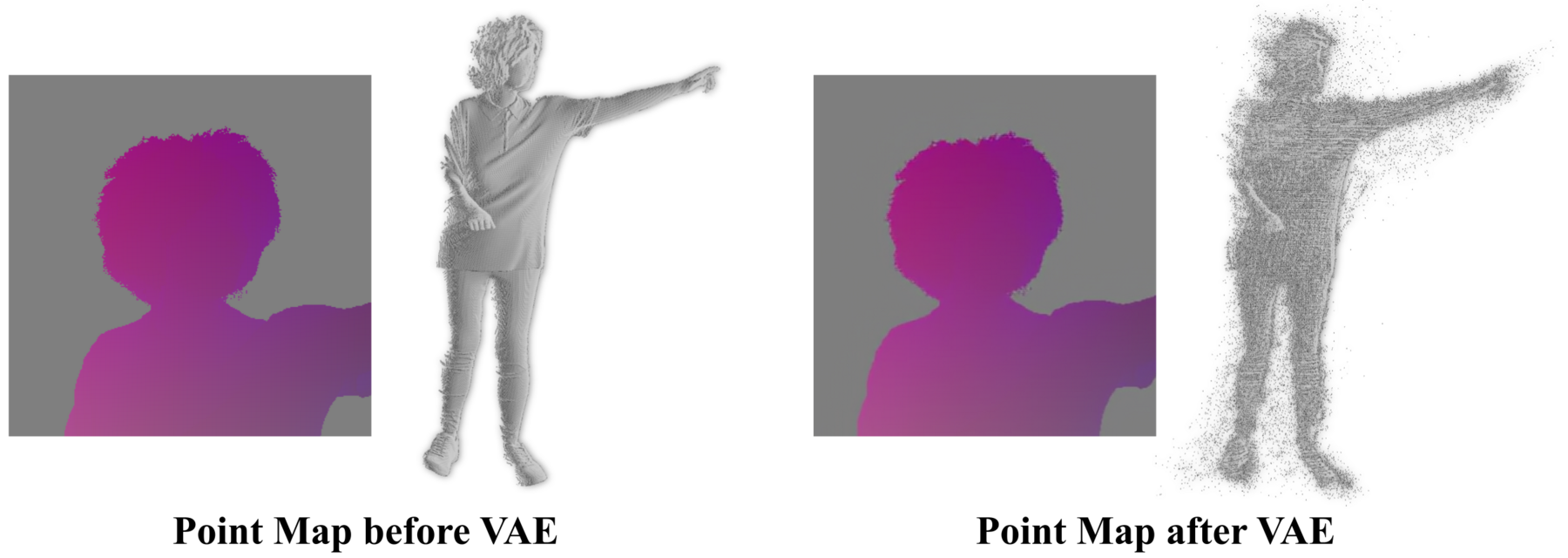}
    \caption{ \textbf{Limitation of Point Map in Latent Space.} As a popular geometry representation~\cite{wang2024dust3r, wang2025vggt} in image sapce, point map shows strong limitation in latent space. Although visually the point map looks similar before and after VAE, the boundary lost huge precision (please zoom in) and it contains much noise after VAE. }
    \label{fig:supp_pointmap}
\end{figure}

Our novel geometry representation, isotropic Normalized Orthographic Depth (iNOD), is a core contribution of our work. Unlike point map which is unsuitable for latent diffusion due to the noise introduced after encoding and decoding (Fig.~\ref{fig:supp_pointmap}), our iNOD is naturally suitable for being applied in latent diffusion model.  We provide a more detailed algorithm for its creation and a justification for our dilation-based boundary refinement.

\subsection{iNOD Creation Algorithm}

\begin{algorithm}
\caption{iNOD Creation from a Metric Depth Map}
\label{alg:inod_creation}
\begin{algorithmic}[1] %
  \State \textbf{Input:} \code{metric_depth} (H, W), \code{cam_intrinsics} (K)
  \State \textbf{Output:} \code{inod_map} (H, W)
  \State
  \State {Step 1: Unproject to 3D point cloud}
  \State \code{pixels_uv} $\gets$ \code{create_pixel_grid}(H, W)
  \State \code{cam_points} $\gets$ \code{unproject}(\code{pixels_uv}, \code{metric_depth}, \code{K})
    \State
  \State {Step 2: Find scaling factor for isotropic normalization}
  \State \code{min_bound} $\gets$ \code{min_axis}(\code{cam_points.xyz})
  \State \code{max_bound} $\gets$ \code{max_axis}(\code{cam_points.xyz})
  \State \code{max_edge} $\gets$ \code{max}(\code{max_bound} - \code{min_bound})

  \State
  \State {Step 3: Apply isotropic normalization}
  \State \code{center} $\gets$ (\code{min_bound} + \code{max_bound}) / 2
  \State \code{norm_points} $\gets$ (\code{cam_points.xyz} - \code{center}) / \code{max_edge}

  \State
  \State {Step 4: Create map via orthographic projection}
  \State \code{iNOD} $\gets$ \code{create_empty_map}(H, W, value = 0)
  \For{$i \gets 1$ to \code{num_points}}
    \State $u, v \gets \code{pixels_uv}[i]$
    \State \code{inod_map}[v, u] $\gets$ \code{norm_points.z}[i]
  \EndFor

  \State
  \State \Return \code{iNOD}
\end{algorithmic}
\end{algorithm}

As described in the main paper (Sec. 3.3), iNOD is generated from a source 3D point cloud. For our \texttt{Synth} data, this is computed from the ground-truth metric depth and known camera intrinsics. The process, detailed in Algorithm 1, is designed to be simple, fast, and distortion-free.

The key steps are (1) Unprojection to a metrically-accurate 3D point cloud, (2) Isotropic Normalization where the entire point cloud is scaled to fit a $[-1, 1]$ cube based on its longest axis, and (3) Orthographic Projection where we simply take the $z$-value of the normalized points and write it into a 2D map. This process preserves the relative 3D geometry and aspect ratio of the subject while ensuring the final 2D map is in a VAE-friendly $[-1, 1]$ range.

\subsection{Dilation for VAE Artifacts}

As noted in the main paper (Sec. 3.3) and visually demonstrated in our main paper's Fig.~2, the pretrained Cosmos VAE~\cite{nvidia2025cosmospredict1}, like most VAEs, can struggle with the extremely sharp boundaries of a foreground silhouette against a blank background.

When a standard iNOD map is passed through the VAE (encoded and then decoded), this compression can introduce minor "ringing" or noise artifacts at the very edge of the geometry. To solve this and ensure a clean, robust geometric boundary after decoding, we apply a simple 2D morphological \textit{dilation} operation to the iNOD map before it is passed to the VAE. We use the assign neighbor iNOD value to dilated pixels. The background pixels are assigned to 0. This slightly thickens the silhouette's boundary, providing a small buffer that is more robust to the VAE's lossy compression. After decoding, we use original mask to "cut off" the dilated region which contains noise. Then we obtain noise-free iNOD geometry.

The effectiveness of this simple step is shown in Fig.2 in main paper. The non-dilated iNOD produces noticeable boundary artifacts after being decoded, while the dilated version results in a clean, sharp, and artifact-free geometric boundary.

\section{Evaluation Details}
This section provides additional specifics on the baseline configurations and metric implementations used for the comparisons in our main paper (Sec 4.3).

To ensure a fair and reproducible comparison, we used publicly available code and pretrained models for all baselines wherever possible. We also tried our best effort for evaluating close-sourced human-centric relighting methods by contacting their authors.  Due to license of our evaluation data~\cite{teufelgera2025HumanOLAT}, we are prohibited to redistribute any of them to conduct quantitative comparison. Hence, we share our in-the-wild evaluation images and focus on qualitative comparison and user study with them.

\subsection{Baseline Configurations}
\label{suppsec:baselines}
\paragraph{Appearance Baselines}
For methods taking square images~\cite{zeng2024dilightnet, jin2024neuralgaffer, zeng2024rgbx} as input, we pad our vertical image horizontally and resize to the desired input resolution. For others we directly resize our image into their desired resolution.

For DiLightNet~\cite{zeng2024dilightnet} which takes additional text prompt as input, we use the unified prompt "A real photo of a human". For IC-Light~\cite{zhang2025iclight} which takes background image instead of environment map as input, we render background image from conditioning environment map using Blender and use as its input.

 For closed-source baselines, we get reply from authors of LuxPostFacto~\cite{mei2025luxpostfacto} and deeply appreciate them for sharing inference results on \textit{face cropped} in-the-wild images with us. LuxPostFacto is a face-centric relighting methods and can have downgrading performance when the input face crop is small or blurry. For TotalRelighting~\cite{pandey2021totalrelighting}, SynthLight~\cite{chaturvedi2025synthlight}, and SwitchLight~\cite{kim2024switchlight}, we cannot get reply from authors on time and will include the comparison at first moment when we obtain their inference results from authors.

\paragraph{Geometry Baselines}
For VGGT~\cite{wang2025vggt}, we use the 1B model and pad our input image into square format and resize to the desired input resolution. For MoGe2~\cite{wang2025moge2} we use the largest model which has 331M parameters and supports metric-scale reconstruction and normal estimation. For Sapiens~\cite{khirodkar2024sapiens}, we use the normal normal estimator with 2B parameters.

\subsection{Metric Implementation Details}
\paragraph{Relighting and Albedo Evaluation} To mitigate the factor of scale ambiguity in lighting and albedo, we follow the protocol from DiffusionRenderer~\cite{liang2025diffusionrenderer} and apply chromatic alignment (a scale-invariant evaluation) between the ground-truth and predicted images for all evaluated methods. All metrics are computed on the foreground pixels.

\paragraph{Geometry (Point Cloud) Evaluation} As described in the main paper, we compare against feed-forward estimators (VGGT, MoGe2) that predict point maps. Our evaluation process is as follows:
\begin{enumerate}
    \item Normalize both the predicted and ground-truth point clouds to fit within  a shared $[-1,1]^3$ bounding box.
    \item Align the predicted shape to the ground-truth shape using the Iterative Closest Point (ICP) algorithm~\cite{arun1987icp}.
    \item Calculate all geometry metrics (CD, F-Score) based on these normalized and aligned shapes.
\end{enumerate}

\paragraph{Normal Evaluation} For surface normal evaluation, all metrics (Angular Error, RMSE) are computed on the foreground pixels using the ground-truth segmentation mask

\section{Additional Results}

\subsection{User Study}

\begin{figure*}
    \centering
    \includegraphics[width=\linewidth]{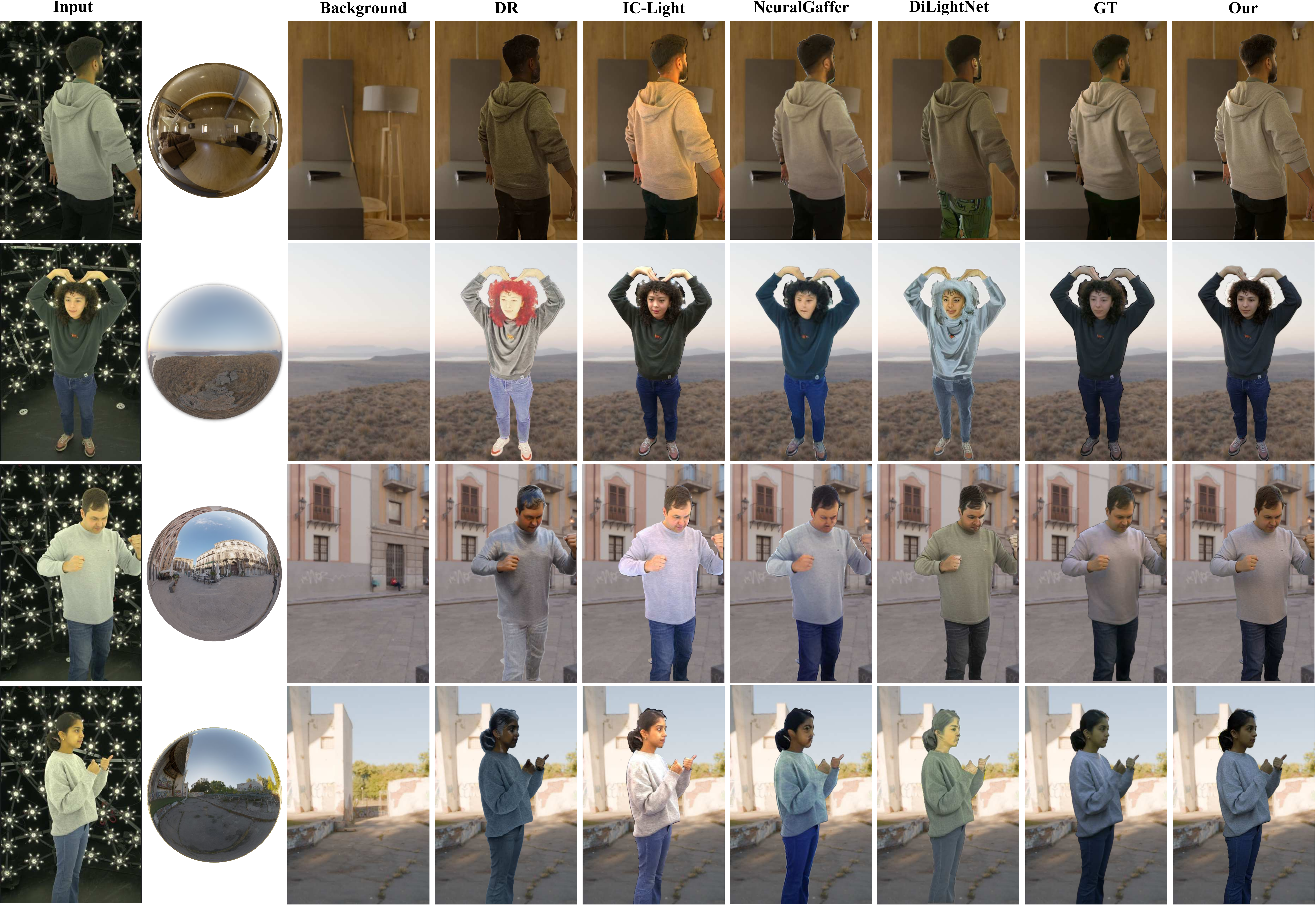}
    \vspace{-2em}
    \caption{\textbf{Qualitative comparison on relighting on HumanOLAT. }
Our model (right) produces more physically-plausible results compared to open-source baselines.}
    \label{fig:supp_relit_comparison_humanolat}
    \vspace{-1em}
\end{figure*}

GeoRelight is capable of jointly generating all five modalities in parallel from a single input image: a high-fidelity relit image under novel illumination, a clean albedo map, a detailed surface normal map, a segmentation mask, and a robust 3D reconstruction. The model generalizes well to out-of-distribution subjects, robustly handling a wide variety of ages, accessories, complex clothing, and non-standard poses. Furthermore, GeoRelight supports controllable generation: rotating the target environment map produces physically-plausible relit images with dynamic shadows, while the estimated albedo and normal maps remain stable, and scaling the light intensity correctly modulates the brightness and shadow depth. We refer readers to our supplementary video for extensive visualization of joint generation results and controllable relighting.

As mentioned in~\ref{suppsec:baselines}, a direct quantitative comparison with closed-source methods like LuxPostFacto~\cite{mei2025luxpostfacto} is not feasible due to data license restrictions. Hence, we conduct a user study to evaluate the performance.

We asked 32 participants to evaluate a set of 20 randomly selected subjects from our in-the-wild test set. The study was divided into two tasks. The first task was a two-alternative forced-choice (2AFC) task for relighting. Participants were shown the input image and environment map, followed by the relit results from our GeoRelight and LuxPostFacto~\cite{mei2025luxpostfacto}, and were asked, "Which relit image is better? Please focus on details such as hair and face". The second task was a three-alternative forced-choice (3AFC) task for geometry. Participants were shown the results from GeoRelight, MoGe2~\cite{wang2025moge2}, and VGGT~\cite{wang2025vggt}. To fairly evaluate the 3D shape, we provided rotating animations of each reconstruction and asked, "Which 3D reconstruction is better? Please focus on details such as eyes and hair". We randomize the order of each method in each question.

We collected a total of 640 votes (32 participants × 20 comparisons) for each task. The results demonstrate an overwhelming preference for our method. For relighting, in the comparison against LuxPostFacto, GeoRelight was preferred in $93.8\%$ (596.5) of votes, while LuxPostFacto was preferred in only $6.2\%$ (43.5) of votes. For geometry, in the reconstruction task, GeoRelight was preferred in $93.4\%$ (598.0) of votes, decisively outperforming both MoGe2 at $5.2\%$ (33.0) and VGGT at $1.4\%$ (9.0).

Our user study measures perceived photorealism and geometric accuracy, which strongly validates our model's superior performance, aligning with the quantitative metrics in the main paper.

\subsection{Qualitative Comparison on Relighting}
In Figure~\ref{fig:supp_relit_comparison_humanolat}, we provide a qualitative comparison for the task of relighting on the HumanOLAT dataset. We compare GeoRelight against state-of-the-art methods, including DiffusionRenderer (DR)~\cite{liang2025diffusionrenderer}, IC-Light~\cite{zhang2025iclight}, NeuralGaffer~\cite{jin2024neuralgaffer}, and DiLightNet~\cite{zeng2024dilightnet}.

Our method consistently demonstrates superior photorealism and physical plausibility. We observe that many baselines struggle with these examples. DiffusionRenderer often produces results that are overly dark or lack contrast. DiLightNet shows significant instability, frequently introducing severe color artifacts. Methods like IC-Light and NeuralGaffer can struggle with correct exposure, color balance, or preserving fine details.

In contrast, GeoRelight produces relit images that are free of artifacts and well-harmonized with the target illumination. Our method properly handles complex interactions between light, hair, and clothing, and accurately renders shadows consistent with the target environment map.

\subsection{Qualitative Comparison on Reconstruction}
We compare our 3D reconstruction against the leading feed-forward geometry estimators, VGGT~\cite{wang2025vggt} and MoGe2~\cite{wang2025moge2}. We also provide qualitative comparison on geometry reconstruction in the main paper (Fig. 7).

We observe that VGGT often produces overly smooth or ``blobby'' reconstructions. It captures the general human form but fails to preserve fine-grained details such as clothing folds, hair, or accessories.

While MoGe2 is able to capture more high-frequency detail, it frequently introduces significant noise, ``floater'' artifacts, and non-manifold holes in the geometry. This results in a brittle and visually unappealing surface.

GeoRelight successfully combines the strengths of both approaches. Our reconstructions are both robustly complete and rich in sharp detail. Our method accurately captures complex surfaces like clothing wrinkles, accessories, and hair, all while maintaining a coherent and clean 3D mesh. Notably, our reconstructions are comparable in quality to methods that use explicit surface representations such as neural surface fields~\cite{xue2023nsf}, while being generated in a single forward pass without iterative optimization. We refer readers to our supplementary video, which shows these 3D reconstructions as rotating animations for a full multi-view comparison.

\section{Extension to Video Relighting}
\label{sec:supp_video}

GeoRelight is primarily designed for single-image relighting and reconstruction by repurposing the temporal dimension $T$ of a pretrained video DiT as a modality dimension $M$. However, the framework can be naturally extended to multi-modal video relighting with minimal architectural modification.

\paragraph{Architecture Modification}
The key change is in the positional encoding. In our static model, we apply a shared 2D RoPE~\cite{su2023rope} ($H{\times}W$) identically across all $M$ modalities to ensure spatial alignment. For the video extension, we replace this with a shared 3D RoPE ($T{\times}H{\times}W$) across all $M$ modalities. This allows the model to reason about both spatial correlations across modalities and temporal consistency across frames, while still maintaining per-modality spatial alignment.

\paragraph{Preliminary Experiment}
To validate this approach, we generated 4,000 synthetic human videos and trained a video variant of GeoRelight. Due to computational constraints, we use 17-frame videos at $1280{\times}832$ resolution, which are compressed to $3{\times}160{\times}104$ latents using the Cosmos $8{\times}8{\times}8$ causal VAE~\cite{nvidia2025cosmospredict1}. The model is initialized from our trained static GeoRelight weights. Our preliminary results show that this video extension achieves meaningful temporal consistency for multi-modal video relighting, confirming that the architecture supports this generalization. A full exploration of video relighting and reconstruction is an exciting direction for future work. Furthermore, our disentangled geometry and appearance outputs could serve as a foundation for downstream tasks such as 3D-consistent avatar creation~\cite{xue2024human3diffusion, xue2025gen3diffusion} and controllable 3D human generation~\cite{xue2025infinihuman}.

\section{Limitations and Failure Cases}
While GeoRelight demonstrates state-of-the-art performance and high generalization, our model shares limitations common to generative frameworks and also has unique failure modes related to its joint-generation task.

\noindent\textbf{Texture-Geometry Ambiguity:} As a single-image model, GeoRelight can struggle with the ambiguity between a 3D object and a 2D texture. For example, a small object held by the subject (e.g., a flower) may be incorrectly interpreted as a 2D texture pattern. In such cases, the object is ``baked'' into the albedo map but is entirely absent from the surface normal and geometry (iNOD) outputs. This leads to a physically implausible relit image where the object appears as a flat pattern rather than a 3D object that should cast its own shadows.

\noindent\textbf{Temporal Inconsistency:} GeoRelight is a single-image framework. Generating a sequence of images (e.g., by rotating the light source) requires multiple independent forward passes. Because the generative process is not deterministic, this can result in small inconsistencies between frames, leading to ``temporal flickering'' when viewed as a video. Enforcing temporal consistency for video relighting is a significant challenge and a clear direction for future work. Alternative sensing modalities such as event cameras, which naturally capture temporal changes with high temporal resolution~\cite{xue2022events, xue2024elnr}, could also provide complementary signals for temporally-consistent reconstruction.

\noindent\textbf{Out-of-Distribution Subjects:} Though our model is robust, it can be challenged by subjects with extreme, out-of-distribution (OOD) poses, severe occlusions, or highly complex, non-Lambertian materials (e.g., metallic or specular clothing) that are rare in our hybrid training data. Our disentangled outputs could also benefit downstream tasks such as physically plausible human-scene interaction estimation~\cite{xue2025physic}, where accurate geometry and lighting are essential for realistic contact and placement.